# Accelerating Partial-Order Planners: Some Techniques for Effective Search Control and Pruning

**Alfonso Gerevini**                                                                   GEREVINI@ING.UNIBS.IT
*Dipartimento di Elettronica per l'Automazione, Università di Brescia*
*Via Branze 38, I-25123 Brescia, Italy*

**Lenhart Schubert**                                                                  SCHUBERT@CS.ROCHESTER.EDU
*Department of Computer Science, University of Rochester*
*Rochester, NY 14627-0226, USA*

## Abstract

We propose some domain-independent techniques for bringing well-founded partial-order planners closer to practicality. The first two techniques are aimed at improving search control while keeping overhead costs low. One is based on a simple adjustment to the default A* heuristic used by UCPOP to select plans for refinement. The other is based on preferring "zero commitment" (forced) plan refinements whenever possible, and using LIFO prioritization otherwise. A more radical technique is the use of operator parameter domains to prune search. These domains are initially computed from the definitions of the operators and the initial and goal conditions, using a polynomial-time algorithm that propagates sets of constants through the operator graph, starting in the initial conditions. During planning, parameter domains can be used to prune nonviable operator instances and to remove spurious clobbering threats. In experiments based on modifications of UCPOP, our improved plan and goal selection strategies gave speedups by factors ranging from 5 to more than 1000 for a variety of problems that are nontrivial for the unmodified version. Crucially, the hardest problems gave the greatest improvements. The pruning technique based on parameter domains often gave speedups by an order of magnitude or more for difficult problems, both with the default UCPOP search strategy and with our improved strategy. The Lisp code for our techniques and for the test problems is provided in on-line appendices.

## 1. Introduction

We are concerned here with improving the performance of "well-founded" domain-independent planners – planners that permit proofs of soundness, completeness, or other desirable theoretical properties. A state-of-the-art example of such a planner is UCPOP (Barrett et al., 1994; Penberthy & Weld, 1992), whose intellectual ancestry includes STRIPS (Fikes & Nilsson, 1971), TWEAK (Chapman, 1987), and SNLP (McAllester & Rosenblitt, 1991). Such planners unfortunately do not perform well at present, in comparison with more practically oriented planners such as SIPE (Wilkins, 1988), PRS (Georgeff & Lansky, 1987), or O-Plan (Currie & Tate, 1991).

However, there appear to be ample opportunities for bringing well-founded planners closer to practicality. In the following, we begin by suggesting some improvements to search control in planning, based on more carefully formulated strategies for selecting partial plans for refinement, and for choosing open conditions in a selected partial plan. Our plan-selection strategy uses S+OC – the number of steps in a plan plus the number of open conditions still to be established – as a heuristic measure for UCPOP's A* search of the plan space. (Addition of an attenuated term reflecting the number of threats or "unsafe conditions" UC, such as 0.1UC, is sometimes advantageous.)[1] Our flaw-selection strategy, which we term ZLIFO, prefers "zero commitment" plan refinements to others, and otherwise uses a LIFO (stack) discipline. Zero commitment refinements are logically necessary ones: they either eliminate a plan altogether because it contains an irremediable flaw, or they add a unique step or unique causal link (from the initial state) to establish an open condition that cannot be established in any other way. The strategy is closely related to ones proposed by Peot & Smith (1993) and Joslin & Pollack (1994) but generally appears to perform better than either.

We describe these two classes of techniques in Section 2 below, and in Section 3 we report our experimental results based on slightly modified versions of UCPOP.[2] For the more difficult problems taken from the available UCPOP test suite and elsewhere, we obtain improvements by factors ranging from 5 to more than 1000, with the hardest problems giving the greatest improvements.

We then turn to our proposal for using computed operator parameter domains during planning. In particular, in Section 4 we motivate and describe a method of precomputing parameter domains based on propagating sets of constants forward from the initial conditions.[3] The process is iterative, but the algorithm runs within a time bound that is polynomial in the size of the problem specification. We provide details of the algorithm, along with theorems about its correctness and tractability, in Sections 4.2–4.3 and Online Appendix 1.

In Section 5 we show how to use parameter domain information in a UCPOP-style planner. During planning, parameter domains can be used to prune operator instances whose parameter domains are inconsistent with binding constraints, and to eliminate spurious threats that cannot, in fact, be realized without violating domain constraints. We illustrate the effectiveness of this technique with examples drawn from the UCPOP test suite as well as from the TRAINS transportation planning world developed at Rochester (Allen & Schubert, 1991; Allen et al., 1995). In some of these tests, we apply the parameter domain information in the context of the default UCPOP search strategy. We demonstrate significant gains on most problems, particularly the more challenging ones (e.g., speedups of more than an order of magnitude for several problems in the STRIPS world, and a more than 900-fold speedup for a TRAINS problem).

In another set of tests in the TRAINS world, we use our own improved search strategies as baseline, i.e., we ask whether *additional* speedups are obtainable by use of parameter domains, above those obtainable with the S+OC and ZLIFO search strategies. Our experimental results again show speedups by about a factor of 10 through use of parameter domains, on top of those obtained by the improved search strategies (the combined speedup is over 2000).

---

1. The search strategy is described as "A* or IDA*" search in (Penberthy & Weld, 1992); in the code for UCPOP 2.0 it is described more generally as best-first, since arbitrary ranking functions, not necessarily corresponding to A* heuristics, may be plugged in. But with choices like S+OC or S+OC+UC as plan-ranking heuristic (as discussed in Section 2.2), it is natural to view the strategy as an A* strategy.
2. While the techniques we describe are applicable to other planners, our focus is on UCPOP because it is well-known and the Lisp code is readily available. The system can be obtained via anonymous ftp from cs.washington.edu.
3. We hope that the notion of a *parameter* domain, as a set of admissible bindings (constants), will cause no confusion with the notion of a *planning* domain, as a specified set of operators, along with constraints on admissible initial conditions and goal conditions.

As evidence that the effectiveness of using parameter domains in combination with our search strategy is not dependent on some peculiarity of the latter, we also include some results for UCPOP's default strategy, Joslin and Pollack's "least cost flaw repair" (LCFR) strategy (Joslin & Pollack, 1994) and for Peot and Smith's "least commitment" (LC) open condition selection strategy (Peot & Smith, 1993) in Section 5.

In Section 6, we state our conclusions, comment on some related work and mention possible extensions of our techniques.

## 2. Plan Selection and Goal Selection

We will be basing our discussion and experiments on UCPOP, an algorithm exemplifying the state of the art in well-founded partial-order planning. Thus we begin with a sketch of this algorithm, referring the reader to (Barrett et al., 1994; Penberthy & Weld, 1992) for details. In the next two subsections we then motivate and describe our improved plan-selection and goal-selection strategies.

### 2.1 UCPOP

UCPOP uses STRIPS-like operators, with positive or negative preconditions and positive or negative effects. The initial state consists of positive predications with constant arguments (if any), and all other ground predications are false by default. Unlike STRIPS, UCPOP also allows *conditional* effects, expressed by 2-part *when*-clauses specifying a (possibly complex) extra condition needed by that effect and the (possibly complex) effect itself. For instance, an action PUTON(?x ?y ?z) ("put ?x on ?y from ?z") might have conditional effects stating that when ?y is not the table, it will not be clear at the end of the action, and when z is not the table, it will be clear at the end of the action. The "U" in UCPOP indicates that universally quantified conditions and effects are permitted as well. For instance, it is permissible to have a precondition for a PICKUP(?x) action that says that for all ?y, (not (on ?y ?x)) holds. Universal statements are handled by explicit substitution of domain constants and need not concern us at this point.

In essence, UCPOP explores a space of partially specified plans, each paired with an agenda of goals still to be satisfied and threats still to be averted. The initial plan contains a dummy *start* action whose effects are the given initial conditions, and a dummy *end* action whose preconditions are the given goals. Thus goals are uniformly viewed as action preconditions, and are uniformly achieved through the effects of actions, including the *start* action.

The plans themselves consist of a collection of *steps* (i.e., actions obtained by instantiating the available operators), along with a set of *causal links*, a set of *binding constraints*, and a set of *ordering constraints*. When an open goal (precondition) is selected from the agenda, it is established (if possible) either by adding a step with an effect that unifies with the goal, or by using an existing step with an effect that unifies with the goal. (In the latter case, it must be consistent with current ordering constraints to place the existing step before the goal, i.e., before the step whose preconditions generated the goal.) When a new or existing step is used to establish a goal in this way, there are several side effects:

- A causal link $(S_p, Q, S_c)$ is also added, where $S_p$ indicates the step "producing" the goal condition $Q$ and $S_c$ indicates the step "consuming" $Q$. This causal link serves to protect the intended effect of the added (or reused) step from interference by other steps.

- Binding constraints are added, corresponding to the unifier for the action effect in question and the goal (precondition) it achieves.

- An ordering constraint is added, placing the step in question before the step whose precondition it achieves.

- If the action in question is new, its preconditions are added to the agenda as new goals (except that EQ/NEQ conditions are integrated into the binding constraints – see below).

- New threats (unsafe conditions) are determined. For a new step and its causal link, other steps threaten the causal link if they have effects unifiable with the condition protected by the causal link (and these effects can occur temporally during the causal link); and the effects of the new step may similarly threaten other causal links. In either case, new threats are placed on the agenda. It is useful to distinguish *definite* threats from *potential* threats: the former are those in which the unification that confirmed the threat involved no new binding of variables.

Binding constraints assert the identity (EQ) or nonidentity (NEQ) of two variables or a variable and a constant. EQ-constraints arise from unifying open goals with action effects, and NEQ-constraints arise (i) from NEQ-preconditions of newly instantiated actions, (ii) from matching negative goals containing variables to the initial state, and (iii) from averting threats by "separation", i.e., forcing non-equality of two variables or a variable and a constant that were unified in threat detection. NEQ-constraints may be disjunctive, but are handled simply by generating separate plans for each disjunct.

The overall control loop of ucpop consists of selecting a plan from the current list of plans (initially the single plan based on *start* and *end*), selecting a goal or threat from its agenda, and replacing the plan by the corresponding refined plans. If the agenda item is a goal, the refined plans are those corresponding to all ways of establishing the goal using a new or existing step. If the agenda item is a *definite* threat to a causal link $(S_p, Q, S_c)$, then there are at most three refined plans. Two of these constrain the threatening step to be before step $S_p$ (demotion) or after step $S_c$ (promotion), thus averting the threat. A third possibility arises if the effect threatening $(S_p, Q, S_c)$ is a *conditional* effect of the threatening action. Such a conditional threat can be averted by creating a goal denying some precondition needed by the conditional effect.

ucpop has a "delay separation" switch, *d-sep*, and when this is turned on, only definite threats are dealt with. Note that potential threats may become definite as a result of added binding constraints. (They may also "expire" as a result of added binding and ordering constraints, i.e., the threatening effect may no longer unify with the threatened condition or it may be forced to occur before or after the threatened causal link. Expired threats are removed from the agenda when selected.) When `*d-sep*` is off, potential threats as well as definite ones are averted, with separation as an additional method of doing so besides the three methods above.

Inconsistencies in binding constraints and ordering constraints are detected when they first occur (as a result of adding a new constraint) and the corresponding plans are eliminated. Planning fails if no plans remain. The success condition is the creation of a plan with consistent binding and ordering constraints and an empty agenda.

The allowance for conditional effects and universal conditions and effects causes only minor perturbations in the operation of UCPOP. For instance, conditional effects can lead to multiple matches against operators for a given goal, each match generating different preconditions. (Of course, there can be multiple matches even without conditional effects, if some predicates occur more than once in the effects.)

The key issues for us right now are the strategic ones: how plans are selected from the current set of plans (discussed in Section 2.2), and how goals are selected for a given plan (discussed in Section 2.3).

## 2.2 The Trouble with Counting Unsafe Conditions

The choice of the next plan to refine in the UCPOP system is based on an A* best-first search. Recall that A* uses a heuristic estimate $f(p)$ of overall solution cost consisting of a part $g(p) =$ cost of the current partial solution (plan) $p$ and a part $h(p) =$ estimate of the additional cost of the best complete solution that extends $p$. In the current context it is helpful to think of $f(p)$ as a measure of plan *complexity*, i.e., "good" plans are simple (low-complexity) plans.

There are two points of which the reader should be reminded. First, in order for A* to guarantee discovery of an *optimal* plan (i.e., the "admissibility" condition), $h(p)$ should not *over*estimate the remaining solution cost (Nilsson, 1980). Second, if the aim is not necessarily to find an optimal solution but to find a satisfactory solution *quickly*, then $f(p)$ can be augmented to include a term that estimates the remaining cost of *finding* a solution. One common way of doing that is to use a term proportional to h(p) for this as well, i.e., we emphasize the $h$-component of $f$ relative to the $g$-component. This is reasonable to the extent that the plans that are most nearly complete (indicated by a low $h$-value) are likely to take the least effort to complete. Thus we will prefer to pursue a plan $p'$ that seems closer to being complete to a plan $p$ further from completion, even though the *overall* complexity estimate for $p'$ may be greater than for $p$ (Nilsson, 1980) (pages 87–88). Alternatively, we could add a heuristic estimate of the remaining cost of finding a solution to $f(p)$ that is more or less independent of the estimate $h(p)$.

With these considerations in mind, we now evaluate the advisability of including the various terms in UCPOP's function for guiding its A* search, namely

S, OC, CL, and UC,

where S is the number of steps in the partial plan, OC is the number of open conditions (unsatisfied goals and preconditions), CL is the number of causal links, and UC is the number of unsafe conditions (the number of pairs of steps and causal links where the step threatens the causal link). The default combination used by UCPOP is S+OC+UC.[4] This becomes S+OC+UC+F if special open conditions called "facts" are present. These are conditions that are not state-dependent (e.g., a numerical relation like (add-one ?x ?y), or a geometrical one like (loc-in-room ?x ?y ?room)) and are established by Lisp functions (Barrett et al., 1994). Since few of our test problems involved facts, we will not discuss the F term further except to say that we followed the UCPOP default strategy of including this term where it is relevant (see the TileWorld problems in Section 3.2 and also some remarks in Section 5.2 in connection with the parameter-domain experiments).

### 2.2.1 S: THE NUMBER OF STEPS CURRENTLY IN THE PLAN

This can naturally be viewed as comprising $g(p)$, the plan complexity so far. Intuitively, a plan is complex to the extent that it contains many steps. While in some domains we might want to make distinctions among the costs of different kinds of steps, a simple step count seems like a reasonable generic complexity measure.

### 2.2.2 OC: THE NUMBER OF OPEN CONDITIONS

This can be viewed as playing the role of $h(p)$, since each remaining open condition must be established by some step. The catch is that it may be possible to use existing steps in the plan (including *start*, i.e., the initial conditions) to establish remaining open conditions. Thus OC can overestimate the number of steps still to be added, forfeiting admissibility.

Despite this criticism, several considerations favor retention of the OC term. First, a better estimator of residual plan complexity seems hard to come by. Perhaps one could modify OC by discounting open conditions that are matched by existing actions, but this presumes that all such open conditions can actually be achieved by action re-use, which is improbable if there are remaining threats, or remaining goals requiring new steps.[5] Second, the possibility that OC will overestimate the residual plan complexity will rarely be actualized, since typically further steps still need to be added to achieve some of the goals, and those steps will typically introduce further open conditions again requiring new steps. Finally, to the extent that OC does at times overestimate the residual plan complexity, it can be viewed as emphasizing the the $h(p)$ term of $f(p)$, thus promoting faster problem-solving as explained above.

### 2.2.3 CL: THE NUMBER OF CAUSAL LINKS

One might motivate the inclusion of this term by arguing that numerous causal links are indicative of a complex plan. As such, CL appears to be an alternative to step-counting.

---

4. This is in no way the "recommended" strategy. The UCPOP implementation makes available various options for controlling search, to be used at the discretion of experimenters. Our present work has prompted the incorporation of our particular strategies as an option in UCPOP 4.0.

5. Note that threats and remaining goals impose constraints that may not be consistent with seemingly possible instances of action re-use. This is clear enough for threats, which often imply temporal ordering constraints inconsistent with re-use of an action. It is also fairly clear for remaining goals. For instance, in Towers of Hanoi the small disk D1 is initially on the medium disk D2, which in turn is on the big disk D3, and D3 is on peg P1. The goal is to move the tower to the third peg P3, so it seems to UCPOP initially as if (on D1 D2) and (on D2 D3) could be achieved by "re-use" of *start*. However, the third goal (on D3 P3) implies that various actions must be added to the plan which are inconsistent with those two seemingly possible instances of action re-use.

However, note that CL is in general larger than S, since every step of a plan establishes at least one open condition and thus introduces at least one causal link. The larger CL is relative to S, the more subgoals are achieved by action re-use. Hence, if we use CL instead of (or in addition to) S in the $g(p)$ term, we would in effect be saying that achieving multiple subgoals with a single step is undesirable; we would tend to search for ways of achieving multiple goals with multiple steps, even when they can be achieved with a single step. This is clearly not a good idea, and justifies the exclusion of CL from $f(p)$.

### 2.2.4 UC: THE NUMBER OF UNSAFE CONDITIONS

We note first of all that this is clearly not a $g$-measure. While the number of threats will tend to increase if we establish more and more subgoals without curtailing threats, threats as such are not elements of the plan being constructed and so do not contribute to its complexity. In fact, when the plan is done all threats will be gone.

Can UC then be viewed as an $h$-measure? One argument of sorts for the affirmative is the following. Not all partial plans are expandable into complete plans, and a high value of UC makes it more likely that the partial plan contains irresolvable conflicts. If we regard impossible plans as having infinite cost, then inclusion of a term increasing with UC as part of the $h$-measure is reasonable. This carries a serious risk, though, since in the case where the partial plan *does* have a consistent completion (despite a high UC-count), inclusion of such a term can greatly overestimate the residual plan complexity.

Another possible affirmative argument is that *conditional* threats are sometimes resolved by "confrontation", which introduces a new goal denying a condition required for the threatening conditional effect. This new goal may in turn require new steps for its achievement, adding to the plan complexity. However, this link to complexity is very tenuous. In the first place, many of the UCPOP test domains involve no conditional effects, and threat removal by promotion, demotion or separation adds no steps. Even when conditional effects are present, many unconditional as well as conditional threats are averted by these methods.

Furthermore, UC could swamp all other terms since threats may appear and expire in groups of size $O(n)$, where $n$ is the number of steps in the plan. For instance, consider a partial plan that involves moves by a robot R to locations L1, ..., Ln, so that there are $n$ causal links labeled (at R L1), ..., (at R Ln). If a new move to location L is now added, initially with an indefinite point of departure ?x, this produces effects (at R L) and (not (at R ?x)). The latter can threaten all of the above $n$ causal links, at least if the new move is at first temporally unordered relative to the $n$ existing moves. If this new action subsequently happens to be demoted so as to precede the first move (or promoted so as to follow the last), or if ?x becomes bound to a constant distinct from L1, ..., Ln, all $n$ threats expire. Keeping in mind that different steps in a plan may have similar effects, we can see that half of the steps could threaten the causal links of the others. In such a case we could have $O(n^2)$ unsafe conditions, destined to expire as a result of $O(n)$ promotions/demotions. In fact even a single new binding constraint may cause $O(n^2)$ threats to expire. For instance, if there are $n/2$ effects (not (P ?x)) threatening $n/2$ causal links labeled (P ?y), then if binding constraint (NEQ ?x ?y) is added, all $n^2/4$ threats expire. Recall that when expired threats are selected from the agenda by UCPOP, they are recognized as such and discarded without further action.

Our conclusion is that it would be a mistake to include UC in full in a general $h$-measure, though some increasing function of UC that remains small enough not to mask OC may be worth including in $h$.

Finally, can UC be regarded as a measure of the remaining cost of *finding* a solution? Here, similar arguments to those above apply. On the affirmative side, we can argue that a high value of UC indicates that we may be facing a combinatorially explosive, time-consuming search for a set of promotions and demotions that produce a conflict-free step ordering. In other words, a high value of UC may indicate a high residual problem-solving cost. (And at the end of such a search, we may still lack a solution, if no viable step ordering exists.) On the other hand, we have already noted that unsafe conditions include many *possible* conflicts which may expire as a result of subsequent partial ordering choices and variable binding choices not specifically aimed at removing these conflicts. So counting unsafe conditions can arbitrarily overestimate the number of genuine refinement steps, and hence the problem-solving effort, still needed to complete the plan.

So UC is scarcely more trustworthy as a measure of residual planning cost than as a measure of residual plan cost.

Thus we conclude that the most promising general heuristic measure for plan selection is S+OC, possibly augmented with an attenuated form of the UC term that will not dominate the S+OC component. (For instance, one might add a small fraction of the term, such as UC/10, or more subtly – to avoid swamping by a quadratic component – a term proportional to $UC^{.5}$.)

### 2.3 The Goal Selection Strategy

An important opportunity for improving planning performance independently of the domain lies in identifying forced refinements, i.e., refinements that can be made *deterministically*. Specifically, in considering possible refinements of a given partial plan, it makes sense to give top priority to open conditions that cannot be achieved; and then preferring open conditions that can be achieved in only one way – either through addition of an action not yet in the plan, or through a unique match against the initial conditions.

The argument for giving top priority to unachievable goals is just that a plan containing such goals can be eliminated at once. Thus we prevent allocation of effort to the refinement of doomed plans, and to the generation and refinement of their doomed successor plans.

The argument for preferring open conditions that can be achieved in only one way is equally apparent. Since every open condition must eventually be established by *some* action, it follows that if this action is unique, it must be part of every possible completion of the partial plan under consideration. So, adding the action is a "zero-commitment" refinement, involving no choices or guesswork. At the same time, adding *any* refinement in general narrows down the search space by adding binding constraints, ordering constraints and threats, which constrain both existing steps and subsequently added steps. For unique refinements this narrowing-down is monotonic, never needing revocation. For example, suppose some refinement happens to add constraints that eliminate a certain action instance $A$ as a possible way of achieving a certain open condition $C$. If the refinement is unique, then we are assured that no completion of the plan contains $A$ as a way of establishing $C$. If it is not unique, we have no such assurance, since some alternative refinement may be compatible with the use of $A$ to achieve $C$. In short, the zero-commitment strategy cuts down the search space without loss of access to viable solutions.

Peot and Smith (1993) studied the strategy of preferring forced threats to unforced threats, and also used a "least commitment" (LC) strategy for handling open conditions. Least commitment always selects an open condition which generates the fewest refined plans. Thus it *entails* the priorities for unachievable and uniquely achievable goals above (while also entailing a certain prioritization of nonuniquely achievable goals). Joslin and Pollack (1994) studied the uniform application of such a strategy to both threats and open conditions in UCPOP, terming this strategy "least cost flaw repair" (LCFR). Combining this with UCPOP's default plan selection strategy, they obtained significant search reductions (though less significant running time reductions, mainly for implementation reasons, but also because of the intrinsic overhead of computing the "repair costs") for a majority of the problems in the UCPOP test suite.

Joslin & Pollack (1994) and subsequently Srinivasan & Howe (1995) proposed some variants of LCFR designed to reduce the overhead incurred by LCFR for flaw selection. These strategies employ various assumptions about the flaw repair costs, allowing the more arduous forms of cost estimation (requiring look-ahead generation of plans) to be confined to a subset of the flaws in the plan, while for the rest an approximation is used that does not significantly increase the overhead. Both teams obtained quite significant reductions in overhead costs in many cases, e.g., by factors ranging from about 3 to about 20 for the more difficult problems. However, overall performance was sometimes adversely affected. Joslin and Pollack found that their variant (QLCFR) solved fewer problems than LCFR, because of an increase in the number of plans generated in some cases. Each of Srinivasan & Howe's four strategies did slightly better than LCFR in some of their 10 problem domains but significantly worse in others. In terms of plans examined during the search, their best overall strategy, which uses similar action instances for similar flaws, did slightly better on 4 of the domains, slightly worse on 4, and significantly worse on 2 (and in those cases the number of plans examined was also more than a factor of 20 above that of default UCPOP).

In the unmodified form of UCPOP, goals are selected from the agenda according to a LIFO (last-in first-out, i.e., stack) discipline. Based on experience with search processes in AI in general, such a strategy has much to recommend it, as a simple default. In the first place, its overhead cost is low compared to strategies that use heuristic evaluation or lookahead to prioritize goals. As well, it will tend to maintain focus on the achievement of a particular higher-level goal by regression – very much as in Prolog goal chaining – rather than attempting to achieve multiple goals in breadth-first fashion.

Maintaining focus on a single goal should be advantageous at least when some of the goals to be achieved are independent. For instance, suppose that two goals G1 and G2 can both be achieved in various ways, but choosing a particular method of achieving G1 does not rule out any of the methods of achieving G2. Then if we maintain focus on G1 until it is solved, before attempting G2, the total cost of solving both goals will just be the sum of the costs of solving them individually. But if we switch back and forth, and solutions of both goals involve searches that encounter many dead ends, the combined cost can be much larger. This is because we will tend to search any unsolvable subtree of the G1 search tree repeatedly, in combination with various alternatives in the G2 search tree (and vice versa). This argument should still have some validity even if G1 and G2 are not entirely

GEREVINI & SCHUBERTindependent; i.e., as long as G1 gives rise to subproblems that tend to fail in the same way regardless of choices made in the attempt to solve G2 (or vice versa), then shifting attention between G1 and G2 will tend to generate a set of partial plans that unnecessarily "cross-multiplies" alternatives.

We have therefore chosen to stay with UCPOP's LIFO strategy whenever there are no zero commitment choices. This has led to very substantial improvements over LCFR in our experiments.

Thus our strategy, which we term ZLIFO ("zero-commitment last-in first-out"), chooses the next flaw according to the following preferences:

1. a definite threat (`*d-sep*` is turned on), using LIFO to pick among these;

2. an open condition that cannot be established in any way;

3. an open condition that can be resolved in only one way, preferring open conditions that can be established by introducing a new action to those that can be established by using `*start*`;[6]

4. an open condition, using LIFO to pick among these.

Hence the overhead incurred by ZLIFO for flaw selection is limited to the open conditions, and is lower for these than the overhead incurred by LCFR. Furthermore, it can also be significantly lower in practice than the overhead incurred by LC, because testing whether an OC is *not* a zero-commitment choice (i.e., whether it can be established in more than one way) is less expensive than computing the total number of ways to achieve it.

In Online Appendix 1 we give the pseudocode of ZLIFO for the selection of the open condition (preferences 2–4). Very recently this implementation has also been packaged into UCPOP 4.0, a new version of UCPOP which is available by anonymous ftp to cs.washington.edu.

## 3. Experiments Using UCPOP

In order to test our ideas we modified version 2.0 of UCPOP (Barrett et al., 1994), replacing its default plan-selection strategy (S+OC+UC) and goal-selection strategy (LIFO) to incorporate strategies discussed in the previous sections.

We tested the modified planner on several problems in the UCPOP suite, emphasizing those that had proved most challenging for previous strategies, on some artificial problems due to Kambhampati *et al.* (1995), in the TRAINS transportation domain developed in Rochester (Allen & Schubert, 1991; Allen et al., 1995), and in Joslin & Pollack's TileWorld domain (Joslin & Pollack, 1994). We briefly describe the test problems and the platforms and parameter settings we used, and then present the experimental results for our improved search strategies.

---

6. 2. and 3. are zero-commitment choices. In our experiments, which are described in the next section, the sub-preference in 3. gave improvements in the context of Russell's tire changing domain (in particular with Fix3), without significant deterioration of performance in the other domains.

104

### 3.1 Test Problems and Experimental Settings

The UCPOP problems include Towers of Hanoi (T of H), Fixa, Fix3, Fixit, Tower-Invert4, Test-Ferry, and Sussman-Anomaly. In the case of T of H, we added a 3-operator version to the UCPOP single-operator version, since T of H is a particularly hard problem for UCPOP and its difficulty has long been known to be sensitive to the formalization (e.g., (Green, 1969)). Fixa is a problem from Dan Weld's "fridge domain", in which the compressor in the fridge is to be exchanged, requiring unscrewing several screws, stopping the fridge, removing the backplane, and making the exchange. Fix3 is from Stuart Russell's "flat tire domain", where a new wheel is to be mounted and lowered to the ground (the old wheel has been jacked up already and the nuts loosened); this requires unscrewing the nuts holding the old wheel, removing the wheel, putting on the new wheel, screwing on the nuts, jacking down the hub, and tightening the nuts. Fixit is more complicated, as the wheel is not yet jacked up initially and the nuts not yet loosened, the spare tire needs to be inflated, and the jack, wrench and pump all need to be taken out of the trunk and stowed again at the end. Tower-Invert4 is a problem in the blocks world, requiring the topmost block in a stack of four blocks to be made bottom-most. Test-Ferry is a simple problem requiring two cars to be moved from A to B using a one-car ferry, by boarding, sailing, and unboarding for each car.

The artificial problems correspond to two parameter settings for ART-$\#_{est}$-$\#_{clob}$, one of the two artificial domains that served as a testbed for Kambhampati *et al.*'s extensive study of the behavior of various planning strategies as a function of problem parameters (Kambhampati et al., 1995). ART-$\#_{est}$-$\#_{clob}$ provides two layers of 10 operators each, where those in layer 1 achieve the preconditions of those in layer 2, and each operator in layer 2 achieves one of the 10 goals. However, some operators in each layer can establish or clobber the preconditions of their neighbors, and this can force operators to be used in a certain order.

The version of the TRAINS domain that we encoded involves four cities (Avon, Bath, Corning, Dansville) connected by four tracks in a diamond pattern, with a fifth city (Elmira) connected to Corning by a fifth track. The available resources, which are located at various cities, consist of a banana warehouse, an orange warehouse, an orange juice factory, three train engines (not coupled to any cars), 4 boxcars (suitable for transporting oranges or bananas), and a tanker car (suitable for transporting orange juice). Goals are typically to deliver oranges, bananas, or orange juice to some city, requiring engine-car coupling, car loading and unloading, engine driving, and possibly OJ-manufacture.

The TileWorld domain consists of a grid on which holes and tiles are scattered. A given tile may or may not fit into a particular hole. The goals are to fill one or more holes by using three possible actions: picking up a tile, going to an x-y location on the grid, and dropping a tile into a hole. The agent can carry at most four tiles at a time.

Formalizations of these domains in terms of UCPOP's language are provided in Online Appendix 2. The experiments for all problems except Fixit, the TRAINS problems and the TileWorld problems were conducted on a SUN 10 using Lucid Common Lisp 4.0.0, while the rest (Tables X–XI in the next subsection) were conducted on a SUN 20 using Allegro Common Lisp 4.2. Judging from some repeated experiments, we do not think that the differences in the platforms significantly impact performance improvements.[7] Among the search control functions provided by UCPOP, we used the default `bestf-search` when the problem was solvable within the search limit of 40,000 plans generated, while we used the function `id-bf-search` (an implementation of the linear-space best-first search algorithm given by Korf, 1992), when this limit was exceeded.[8] In all of the experiments the delay-separation switch, `*d-sep*`, was on, except for those using the LCFR strategy.

| Goal-selection | Plan-selection | CPU sec | Plans |
|---|---|---|---|
| LIFO | S+OC+UC | 204.51 | 160,911/107,649 |
| LIFO | S+OC | 0.97 | 751/511 |
| ZLIFO | S+OC+UC | 6.90 | 1816/1291 |
| ZLIFO | S+OC | 0.54 | 253/184 |

Table I: Performance of plan/goal selection strategies on T-of-H1

### 3.2 Experimental Results for ZLIFO and S+OC

Tables I–XI show the CPU time (seconds) and the number of plans created/explored by UCPOP on twelve problems in the domains described above: Towers of Hanoi with three disks and either one operator (T-of-H1) or three operators (T-of-H3), the fridge domain (Fixa), the tire changing domain (Fix3 and Fixit), the blocks world (Tower-Invert4 and Sussman-anomaly), the ferry domain (Test-Ferry), the artificial domain ART-$\#_{est}$-$\#_{clob}$ (specifically, ART-3-6 and ART-6-3), the TRAINS domain (Trains1, Trains2 and Trains3) and the TileWorld domain (tw-1, ..., tw-6). Both the number of plans created/explored and the CPU time are important performance measures. The number of plans, which indicates search space size, is a more stable measure in the sense that it depends only on the search algorithm, not the implementation.[9] But the time is still of interest since an improvement in search may have been purchased at the price of a more time-consuming evaluation of alternatives. It turns out that we do pay some price in overhead when we substitute our strategies for the defaults (factors ranging from about 1.2 to 1.9, and rarely higher, per plan created). This may be due to slightly greater inherent complexity of ZLIFO versus LIFO, but we think the differences could be reduced by substituting modified data structures for those of UCPOP – we were committed to not altering these.

Tables I and II show that for the T of H the plan selection strategy S+OC gives dramatic improvements over the default S+OC+UC strategy. (In these tests the default LIFO goal selection strategy was used.) In fact, UCPOP solved T-of-H1 in 0.97 seconds using S+OC versus 204.5 seconds using S+OC+UC. T-of-H3 proved harder to solve than T-of-H1, re-quiring 8.5 seconds using S+OC and an unknown time in excess of 600 CPU seconds using S+OC+UC.

---

7. The differences were the result of what was available at different times and locales over the course of nearly two years of experimentation.
8. This choice was motivated by the observation that when the problem is relatively easy to solve `bestf-search` appears to be more efficient than `id-bf-search`, while for hard problems it can be very inefficient because of the considerable amount of space used at run time and the CPU time spent on garbage collection, which in some cases made Lisp crash, reporting an internal error.
9. It is also worth noting that the number of plans created implicitly takes into account plan size, since addition of a step to a plan is counted as creation of a new plan in UCPOP.

| Goal-selection | Plan-selection | CPU sec | Plans |
|---|---|---|---|
| LIFO | S+OC+UC | > 600 | > 500,000 |
| LIFO | S+OC | 8.54 | 5506/3415 |
| ZLIFO | S+OC+UC | > 600 | > 500,000 |
| ZLIFO | S+OC | 1.24 | 641/420 |

Table II: Performance of plan/goal selection strategies on T-of-H3

| Goal-selection | Plan-selection | CPU sec | Plans |
|---|---|---|---|
| LIFO | S+OC+UC | 2.45 | 2131/1903 |
| LIFO | S+OC | 2.48 | 2131/1903 |
| ZLIFO | S+OC+UC | 0.33 | 96/74 |
| ZLIFO | S+OC | 0.33 | 96/74 |

Table III: Performance of plan/goal selection strategies on Fixa

Our ZLIFO goal-selection strategy can significantly accelerate planning compared with the simple LIFO strategy. In particular, when ZLIFO was combined with the S+OC plan-selection strategy in solving T of H, it further reduced the number of plans generated by a factor of 3 in T-of-H1 and by a factor of 8 in T-of-H3. The overall performance improvement for T-of-H1 was thus a factor of 636 in terms of plans created and factor of 379 in terms of CPU time (from 204.5 to 0.54 seconds).

Tables III–VIII provide data for problems that are easier than T of H, but still challenging to UCPOP operating with its default strategy, namely Fixa (Table III), Fix3 (Table IV), Tower-Invert4 (Table V), Test-Ferry (Table VI) and the artificial domain ART-$\#_{est}$-$\#_{clob}$ with $\#_{est} = 3$ and $\#_{clob} = 6$ (Table VII) and with $\#_{est} = 6$ and $\#_{clob} = 3$ (Table VII). The results show that the combination of S+OC and ZLIFO substantially improves the performance of UCPOP in comparison with its performance using S+OC+UC and LIFO. The number of plans generated dropped by a factor of 22 for Fixa, by a factor of 5.9 for

| Goal-selection | Plan-selection | CPU sec | Plans |
|---|---|---|---|
| LIFO | S+OC+UC | 6.50 | 3396/2071 |
| LIFO | S+OC | 0.43 | 351/215 |
| ZLIFO | S+OC+UC | 1.12 | 357/221 |
| ZLIFO | S+OC | 1.53 | 574/373 |

Table IV: Performance of plan/goal selection strategies on Fix3

107

| Goal-selection | Plan-selection | CPU sec | Plans |
|---|---|---|---|
| LIFO | S+OC+UC | 1.35 | 808/540 |
| LIFO | S+OC | 0.19 | 148/105 |
| ZLIFO | S+OC+UC | 2.81 | 571/378 |
| ZLIFO | S+OC | 0.36 | 142/96 |

Table V: Performance of plan/goal selection strategies on Tower-Invert4

| Goal-selection | Plan-selection | CPU sec | Plans |
|---|---|---|---|
| LIFO | S+OC+UC | 0.63 | 718/457 |
| LIFO | S+OC | 0.32 | 441/301 |
| ZLIFO | S+OC+UC | 0.24 | 136/91 |
| ZLIFO | S+OC | 0.22 | 140/93 |

Table VI: Performance of plan/goal selection strategies on Test-Ferry

| Goal-selection | Plan-selection | CPU sec | Plans |
|---|---|---|---|
| LIFO | S+OC+UC | .67 | 568/392 |
| LIFO | S+OC | 1.36 | 1299/840 |
| ZLIFO | S+OC+UC | 0.16 | 72/49 |
| ZLIFO | S+OC | 0.18 | 79/54 |

Table VII: Performance of plan/goal selection strategies on ART-$\#_{est}$-$\#_{clob}$ with $\#_{est} = 3$ and $\#_{clob} = 6$ (averaged over 100 problems)

| Goal-selection | Plan-selection | CPU sec | Plans |
|---|---|---|---|
| LIFO | S+OC+UC | 1.32 | 985/653 |
| LIFO | S+OC | 2.08 | 1743/1043 |
| ZLIFO | S+OC+UC | 0.14 | 57/37 |
| ZLIFO | S+OC | 0.14 | 57/37 |

Table VIII: Performance of plan/goal selection strategies on ART-$\#_{est}$-$\#_{clob}$ with $\#_{est} = 6$ and $\#_{clob} = 3$ (averaged over 100 problems)

| Goal-selection | Plan-selection | CPU sec | Plans |
|---|---|---|---|
| LIFO | S+OC+UC | 0.06 | 44/26 |
| LIFO | S+OC | 0.04 | 36/21 |
| ZLIFO | S+OC+UC | 0.12 | 67/43 |
| ZLIFO | S+OC | 0.07 | 41/25 |

Table IX: Performance of plan/goal selection strategies on Sussman-anomaly

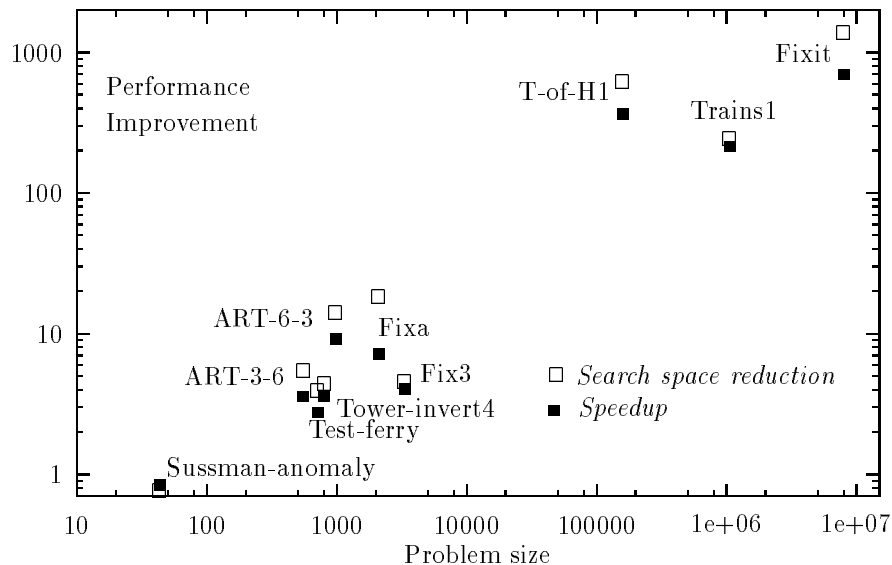

Figure 1: Performance improvement due to ZLIFO and S+OC, relative to the number of plans generated by LIFO and S+OC+UC (log-log scale). The improvements for the problems that UCPOP was unable to solve even with a very high search limit (Trains2, Trains3, and T-of-H3) are not included.

Fix3, by a factor of 5.7 for Tower-Invert4, by a factor of 5.1 for Test-Ferry, by a factor of 7 for ART-3-6, and by a factor of 17 for ART-6-3.

Concerning ART-$\#_{est}$-$\#_{clob}$, note that the performance we obtained with unenhanced UCPOP (568 plans generated for ART-3-6 and 985 for ART-6-3) was much the same as (just marginally better than) reported by Kambhampati et al. (1995) for the best planners considered there (700 − 1500 plans generated for ART-3-6, and 1000-2000 for ART-6-3). This is to be expected, since UCPOP is a generalization of the earlier partial-order planners. Relative to standard UCPOP and its predecessors, our "accelerated" planner is thus an order of magnitude faster. Interestingly, the entire improvement here can be ascribed to ZLIFO (rather than S+OC plan selection, which is actually a little worse than S+OC+UC). This is probably due to the unusual arrangement of operators in ART-$\#_{est}$-$\#_{clob}$ into a "clobbering chain" ($A_{n_-,1}$ clobbers $A_{n_--1,1}$'s preconditions, ..., $A_{1,1}$ clobbers $A_{0,1}$'s preconditions; similarly for $A_{i,2}$), which makes immediate attention to new unsafe conditions an unusually good strategy.

In experimenting with various combinatorially trivial problems that unmodified UCPOP handles with ease, we found that the S+OC and ZLIFO strategy is neither beneficial nor harmful in general; there may be a slight improvement or a slight degradation in performance. Results for the Sussman anomaly in Table IX provide an illustrative example.

We summarize the results of Tables I–X in Figure 1, showing the performance improvements obtained with the combined ZLIFO goal selection strategy and S+OC plan selection strategy *as a function of problem difficulty* (as indicated by the number of plans generated by the default LIFO plus S+OC+UC strategy). The trend toward greater speedups for more complex problems (though somewhat dependent on problem type) is quite apparent from the log-log plot.

|  |  | Trains1 | Trains2 | Trains3 | Fixit |
|---|---|---|---|---|---|
| **ZLIFO &** | Plans | 4097/2019 | 17,482/10,907 | 31,957/19,282 | 5885/3685 |
| **S+OC** | Time | 13.7 | 80.6 | 189.8 | 32.5 |
| **LC &** | Plans | 438/242 | 34,805/24,000 | 253,861/168,852 | 71,154/46,791 |
| **S+OC** | Time | 2.6 | 368.9 | 1879.9 | 547.8 |
| **LCFR &** | Plans | 1093/597 | >1,000,000 | >1,000,000 | 190,095/117,914 |
| **S+OC** | Time | 10.65 | >10,905 | >9918 | 4412.36 |
| **LIFO &** | Plans | 1,071,479/432,881 | > 10,000,000 | > 1,000,000 | 8,090,014/4,436,204 |
| **S+OC+UC** | Time | 3050.15 | > 37,879 | > 2539 | 27,584.9 |

Table X: Performance of the plan selection strategy S+OC in combination with the goal selection strategies ZLIFO, LCFR and LC in solving problems which are very hard for the default strategies of UCPOP (S+OC+UC/LIFO). (The CPU seconds do not include Lisp garbage collection. The number of plans generated for LCFR does not include those created in order to estimate the repair cost of the flaws.)

| Problem | ZLIFO* | | LCFR | |
|---|---|---|---|---|
| name | CPU time | Plans | CPU time | Plans |
| tw-1 | 0.09 | 26/15 | 0.10 | 26/15 |
| tw-2 | 0.61 | 72/39 | 0.66 | 72/39 |
| tw-3 | 2.55 | 138/71 | 3.17 | 139/72 |
| tw-4 | 7.80 | 224/111 | 10.97 | 227/114 |
| tw-5 | 19.41 | 330/159 | 30.17 | 336/165 |
| tw-6 | 42.57 | 456/215 | 71.10 | 466/225 |

Table XI: Performance of UCPOP in the TileWorld domain using ZLIFO* and LCFR for goal selection, and S+OC+F+0.1UC for plan selection

For direct comparison with Joslin and Pollack's LCFR strategy and Peot and Smith's LC strategy, we implemented their strategies and applied them to several problems. They did very well (sometimes better than ZLIFO) for problems on the lower end of the difficulty spectrum, but poorly for harder problems. (For all the problems we ran, LC with the `*d-sep*` switch on performed better than LCFR in terms of plans explored and CPU time required.) For T-of-H1 LCFR in combination both with the default S+OC+UC plan selection strategy, and with our S+OC plan strategy did not find a solution within a search limit of 200,000 plans generated (cf. 253 for ZLIFO with S+OC, and 751 for ZLIFO with S+OC+UC), requiring an unknown CPU time in excess of 4254 seconds with S+OC+UC, and in excess of 4834 seconds with S+OC (cf. 0.54 seconds for ZLIFO with S+OC).[10] LC performed much better than LCFR but still considerably worse than ZLIFO, solving T-of-H1 by generating/exploring 8313/6874 plans with S+OC and 8699/6441 plans with S+OC+UC, and requiring 44.4 CPU secs. and 48.95 CPU secs. respectively. For T-of-H3, LC found a solution by generating/exploring 21,429/15,199 plans with S+OC+UC and 17,539/14,419 plans with S+OC, requiring 145.18 CPU secs. and 77.84 CPU secs. respectively.

Table X shows the results for the plan strategy S+OC, with the goal strategies ZLIFO, LCFR and LC, applied to three problems (Trains1, Trains2 and Fixit). As shown by the data in the table these are very hard for the default strategies of UCPOP (LIFO & S+OC+UC), but become relatively easy when S+OC is used in combination either with ZLIFO, LCFR or LC. While LCFR and LC did slightly better than ZLIFO for Trains1 (the easiest of these problems), they performed quite poorly for Fixit, Trains2 and Trains3 (the hardest problems) compared to ZLIFO.

Joslin and Pollack (1994) tested their LCFR strategy on six problems in the TileWorld (tw-1, ..., tw-6), five of which are very hard for default UCPOP, but easy for UCPOP using LCFR.[11] We tested our ZLIFO strategy in the TileWorld using the same six problems. ZLIFO did well for tw-1–4, but for tw-5 and tw-6 its performance dropped well below that of LCFR. This raised the question whether for these particular problems it is crucial to minimize "repair cost" in flaw selection uniformly, rather than just in certain special cases (ZLIFO does minimize the repair cost when no threat is on the flaw list, and at least one zero-commitment open condition is present). However, further experiments aimed at answering this question suggested that the poor choices made by ZLIFO for some TileWorld problems were not due to selection of "high cost" over "low cost" flaws. Instead two factors appear be crucial for improving ZLIFO: (a) emphasizing zero-commitment open conditions by giving them higher priority than threats; (b) when there are no zero-commitment open conditions, resolving threats as soon as they enter the agenda. (We realized the relevance of (b) by observing that the performance of a modified versions of LCFR, where the `*d-sep*` switch is implicitly forced on, dramatically degraded for tw-6 in a slightly different formulation of the TileWorld.)

We extended our ZLIFO strategy to include (a) and (b), and we briefly tested the resulting variant of ZLIFO (ZLIFO*). Table XI shows the results for ZLIFO* together with the plan selection strategy S+OC+0.1UC+F, where as discussed in Section 2.3 we included an attenuated form of the UC term (UC/10), and an F term equal to the number of facts since TileWorld uses facts to track the number of tiles carried by the agent.[12] ZLIFO* performed very efficiently for all six TileWorld problems, in fact a little better than LCFR. Note that for these problems ZLIFO* is more efficient than LCFR in terms of the CPU time, even though the number of plans generated/explored by the two strategies is approximately the same. This is because the overhead of selecting the next flaw to be handled is higher in LCFR than in ZLIFO* (and ZLIFO). In fact, while LCFR needs to compute the "repair cost" of *each* flaw (including the threats) in the current plan, ZLIFO* (ZLIFO) only needs to check for the presence of zero-commitment open conditions, without processing the threats.

---

10. This was with `*d-sep*` turned off, which is the implicit setting in LCFR (Joslin, 1995). In our experiments we also tested a variant of LCFR, where the switch is forced to be on. The resulting goal strategy in combination with our plan strategy S+OC performed significantly better for T-of-H1, solving the problem generating/exploring 7423/6065 plans, and using 110.45 CPU seconds. Note also that a comparison of our implementation of LCFR and Joslin & Pollack's implementation used for the experiments discussed in (Joslin & Pollack, 1994) showed that our implementation is considerably faster (Joslin, 1995).
11. In their experiments tw-2, the easiest among tw-2–6, was not solved by UCPOP even when allowed to run for over eight hours. On the other hand, UCPOP using LCFR solves tw-6, the hardest problem, without ever reaching a dead-end node in the search tree.
12. In the ZLIFO* experiments the refined plans generated by resolving a threat were added to the flaw list in the following order: first the plan generated by promotion, then the plan generated by demotion, and finally the plan generated by confrontation or separation.

Additional experiments indicated that the average performance of ZLIFO* is comparable to that of ZLIFO for most of the other problems we used in our experiments, in terms of plans created/explored. However, the CPU time tends to increase since the overhead of computing the goal selection function is higher for ZLIFO* than for ZLIFO, because of the extra agenda-management costs. Because of this overhead, we do not regard ZLIFO* as generally preferable to ZLIFO. However, the TileWorld experiments underscored for us that in some worlds refinements of ZLIFO are advantageous.

Finally, another possible variant of ZLIFO, which was suggested to us by David Smith, is based on the following preferences of the next flaw to be handled: (i) a threat that cannot be resolved; (ii) an open condition that cannot be established; (iii) a threat that has only one possible resolution; (iv) an open condition that can only be established in one way; (v) other threats; (vi) other open conditions (using LIFO to pick among these). We observe that while this strategy could give further savings in terms of plans created/explored, it also imposes an additional overhead with respect to both ZLIFO and ZLIFO* which could degrade performance in terms of CPU time.

## 4. Precomputing Parameter Domains

Even with the speedups obtained through improved search, a UCPOP-like algorithm remains severely limited in the complexity of problems it can solve. We believe that significant further progress requires fuller use of *global* properties of the search space, as determined by the structure of the operators, initial conditions, and goals. One way to do that would be through a more in-depth analysis of alternatives during the search, but this can lead to high overhead costs. Another is to *precompute* constraints on the search space, and to use these during planning to prune the search. The parameter domain method we now motivate and describe is of the latter type.

### 4.1 How Can Parameter Domains Help?

In our previous experimentation with UCPOP strategies, we found that UCPOP goal regression often hypothesized steps that were doomed to be abandoned eventually, because they stipulated impossible parameter bindings. A clear example of this occurred in the Molgen domain, as encoded in the UCPOP test suite. The goal of the "Rat-insulin" test problem is

```
(and (bacterium ?b) (molecule ?m)
     (contains IG ?m) (contains ?m ?b) (pure ?b)),
```

where ?b and ?m are existentially quantified variables. What this means is that we wish to create a purified bacterial culture ?b, where ?b contains a molecule ?m (necessarily an exosome, it turns out), and this molecule in turn contains the insulin gene, IG. We are using the abbreviations IG, EE, JE, L for insulin-gene, e-coli-exosome, junk-exosome, and linker; and E, J, A1 for e-coli, junk, and antibiotic-1. Roughly speaking, the solution involves processing the initially given mRNA form of the insulin gene so as to produce a form of insulin DNA that can be spliced into the e-coli-exosome, using a ligate operator. In turn, the exosome is inserted into the e-coli bacterium using a transform operator, and the bacterial culture is then purified using a screen operator, with antibiotic-1. (The junk bacterium and exosome merely serve to complicate the task – they are nearly, but not quite, substitutable for the e-coli bacterium and exosome; the junk exosome, unlike e-coli-exosome, is not resistant to antibiotic-1, violating a precondition of screen.)

Now, in the initial regression the goals (bacterium ?b) and (molecule ?m) can be established only with the *start* operator, i.e., with the initial conditions, and thus will not be instantiated to bizarre values. (The initial conditions supply E and J as the only instances of bacterium, and IG, EE, JE, and L as the only instances of molecule.) On the other hand, the remaining goals turn out to match the effects of various instances of the ligate, transform, and screen operators of Molgen, as follows:

```
(contains IG ?m): (ligate IG ?m), (transform IG ?m)
(contains ?m ?b): (ligate ?m ?b), (transform ?m ?b)
(pure ?b):        (screen ?b ?y ?z)
```

UCPOP will happily regress on these actions. Yet two of them, (transform IG ?m) and (ligate ?m ?b), are doomed to fail, perhaps after a great deal of effort has been expended on trying to satisfy their preconditions. In particular, examination of the constants that can "flow into" the transform operator from the initial conditions and other Molgen operators shows that its first argument is restricted to domain {EE, JE}, i.e., it must be one of the given exosomes, and the second is restricted to {E, J}, i.e., it must be one of the given bacteria. Consequently the instance (transform IG ?m) is unrealizable, as its first argument IG is not in {EE, JE}. (Note that distinct constants denote distinct entities according to the unique-names assumption made by UCPOP.) The (ligate ?m ?b) action is doomed for slightly more subtle reasons. It is the result of a match between (contains ?m ?b) and a "*when*-clause" (conditional effect) of the ligate operator, whose preconditions can be reached only if the second parameter ?b lies in the set of molecules {IG, JE, EE}; yet ?b is also restricted to the set of bacteria {E, J}, as a result of the goal condition (bacterium ?b). The fact that these sets are disjoint should allow us to eliminate the (transform IG ?m) action.

Note that elimination of action candidates as above increases the number of zero commitment plan refinements that can be made. In the example, we are left with exactly one action for each of the three goals, and so the ZLIFO and LCFR strategies will prefer to regress on these goals rather than regressing on (bacterium ?b) and (molecule ?m) – which would prematurely make arbitrary choices of ?b and ?m from the initial state.

### 4.2 Description of the Algorithm

In any completed plan, each precondition of each action must be instantiated by an effect of some earlier action. So the values of the parameters of the action can only be values that can be "produced" by earlier actions, starting with the initial action, *start*. Moreover, suppose that a parameter $x$ of a certain action occurs in each of preconditions *P1, ..., Pk*. Then a constant $c$ is a possible value of $x$ only if earlier actions can instantiate $x$ to $c$ in *each* of *P1, ..., Pk*.

Our algorithm find-parameter-domains is based on these observations. Beginning in the initial state, it propagates positive atomic predications to all possible operator preconditions. For a propagated ground atom, if the atom matches an operator precondition, the algorithm adds the constants in that ground atom to the *individual domains* of the parameters they were unified with. These individual domains are particular to specific preconditions. For instance, the individual domain of ?x for an operator with preconditions (on ?x ?y), (clear ?x) will in general be distinct for these two preconditions.

As soon as we have nonempty individual domains for all parameters in all preconditions of an operator, we form the *intersection* of the individual domains of each parameter of the operator. For example, if (on ?x ?y) has (so far) been matched by (on A B) and (on B C), and (clear ?x) has (so far) been matched by (clear A) and (clear Table), then the individual domain of x will be {A,B} in the first precondition and {A,Table} in the second. Thus (assuming there are no other preconditions) the intersected domain of ?x will be {A} at this point. If later (clear B) is also matched against (clear ?x), the intersected domain of ?x will grow to {A,B}. When both ?x and ?y have nonempty intersected domains, the effects (postconditions) of the operator can in turn be propagated, with ?x and ?y "bound" to their intersected domains.

The propagated effects are again matched against all possible operator preconditions, and when a variable "bound" to an intersected domain is successfully unified with a variable in a precondition, it passes its intersected domain to the *individual* domain of that precondition-variable (via a union operation). This can again lead to growth of the intersected domains of the operator whose precondition was matched, the effects of that operator may then be propagated, and so on. The individual domains and intersected domains grow monotonically during the propagation process, and in the end represent the desired parameter domains of the operators.

We illustrate this process through an example. Consider the simple planning problem depicted in Figure 2 where an "operator graph" (Smith & Peot, 1993) is used to describe the logical dependencies among the operators, while the iterative computation of the parameter domains is graphically illustrated with a "domain-propagation graph" below the operator graph.

The initial conditions (P A) and (P B) unify with the precondition (P ?x) of op1. So, the individual domain of ?x relative to the precondition P of op1 is {A,B}. On the other hand, the precondition (Q ?x) of op1 cannot be satisfied by the initial state, and so the individual domain of ?x relative to Q is initially the empty set. Hence the intersected domain of ?x for op1 is also the empty set.

For op2 we have a different situation, since here we have only one precondition and it can be established by the initial state. Therefore, the individual domain of ?y relative to precondition R of op2 is the set of constants {B,C}, and the intersected domain of ?y for op2 is the same set (because R is the only precondition of op2 involving ?y). Since the intersected domain of ?y has been enlarged (initially it was empty), it is propagated to the individual domains of the other operators through the effect (Q ?y) of op2. In particular,

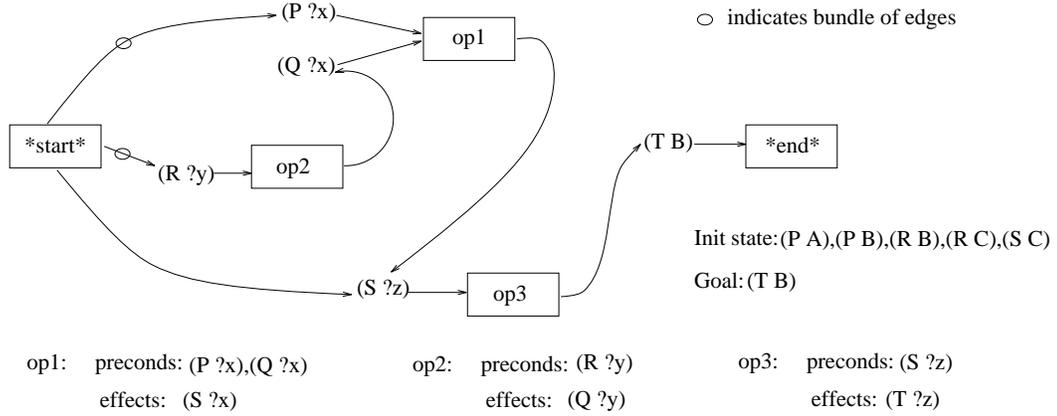

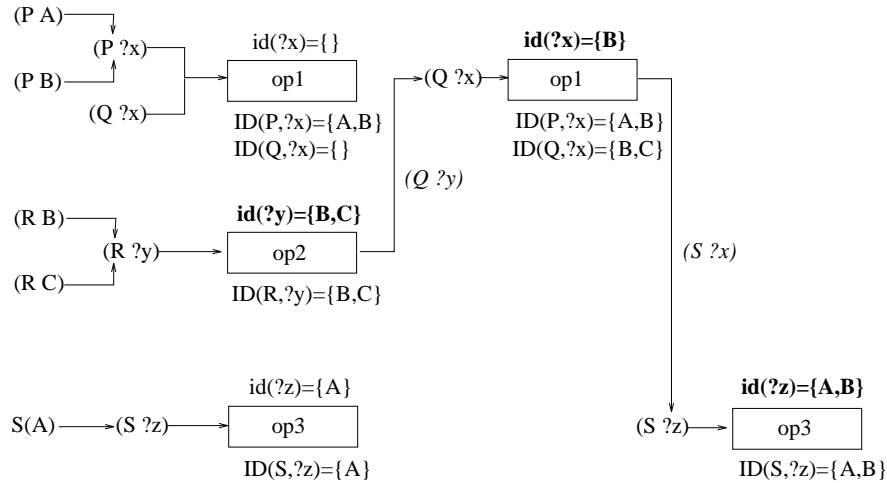

Figure 2: Operator and domain-propagation graphs for a simple planning problem. ID(?x,P) indicates the individual domain of the parameter ?x relative to precondition P; id(?x) indicates the intersected domain of the parameter ?x; final intersected domains are indicated using bold fonts.

(Q ?y) matches the precondition (Q ?x) of op1. So, the individual domain of ?x relative to precondition Q of op1 is updated by adding the constants of the intersected domain of ?y to it. Thus the intersected domain of ?x is enlarged to {B}, and can be propagated through the effect (S ?x) of op1.

Similarly, the propagation of (S ?x) will enlarge the individual domain of ?z for op3, and also the intersected domain, to the set {A,B}. Therefore, the final intersected domains are: {B} for ?x in op1; {B,C} for ?y in op2; {A,B} for ?z in op3.

Before presenting the algorithm a little more formally, we note that the parameter domains will sometimes be "too large", including values that would be found to be impossible if a more detailed state space exploration were conducted. However, all that is required for soundness in our use of the domains is that they not be "too small" (i.e., that they contain all parameter values that can actually occur in the problem under consideration). Of course, to be of practical use the parameter domains of an operator should exclude some of the constants occurring in the problem specification, particularly those for which it is intuitively obvious that they are of the wrong sort to fill particular argument slots of the operator. This has turned out to be the case for all problem domains we have so far experimented with.

The preceding sketch of our method is an oversimplification since preconditions and effects of UCPOP operators may be particular to a *when*-clause. In this case we compute individual domains and intersected domains separately for each *when*-clause. For example, consider the following schematic representation of an operator:

```
(define (operator op1)
   :parameters (?x ?y)
   :precondition (and P1 P2)
   :effect (and E1 E2
             (when P'E')
             (when P"E") )),
```

where all conditions starting with *P* or *E* denote atomic formulas that may involve ?x and ?y. We can think of this operator as consisting of a *primary when-clause* whose preconditions *P1* and *P2* must always be satisfied and whose effects *E1* and *E2* are always asserted, and two *secondary when-clauses* whose respective preconditions *P'* and *P"* may or may not be satisfied, and when they are, the corresponding effects *E'* and *E"* are asserted. Here our algorithm would maintain individual domains for ?x and ?y for each of preconditions *P1, P2, P'*, and *P"*, and it would maintain intersected domains for ?x and ?y for the primary *when*-clause and each of the two secondary clauses. The intersected domains for the secondary clauses would be based on the individual domains of ?x and ?y not only relative to *P'* and *P"*, but also on those relative to *P1* and *P2*, since (as noted) the primary preconditions must hold for the operator to have any of its effects, including conditional effects.

Some further complications arise when UCPOP operators contain universally quantified preconditions or effects, disjunctive preconditions, or facts (mentioned in Section 2.2). Rather than dealing with these complications directly, we will assume that no such operators occur in the input to the algorithm. Later we describe a semi-automated way of handling operators containing the additional constructs.

The algorithm is outlined below (a more detailed description is given in Online Appendix 1). W is a list of (names of) *when*-clauses whose effects are to be propagated. Individual parameter domains are initially nil, and intersected parameter domains are initially either nil or T (where T is the universal domain). The intersected domain of a parameter, relative to a given *when*-clause, is T just in case the parameter occurs neither in the preconditions of the *when*-clause nor in the primary preconditions. (In such a case the successful instantiation of the *when*-clause is clearly independent of the choice of value for the parameter in question.) Unification in step 2(a) is as usual, except that when an effect variable v is unified with a constant c in a precondition, the unification succeeds, with unifier v = c, just in case c is an element of the intersected domain of v (for the relevant *when*-clause). The given *inits* (initial conditions) and *goals* (which may be omitted, i.e., nil) are treated as an operator `*start*` with no preconditions and an operator `*end*` with no effects. Variables in *goals* are treated like operator parameters. We use the terms "parameters" and "variables" interchangeably here.

Algorithm: **find-parameter-domains**(*operators, inits, goals*)

1. Initialize W to the initial conditions, so that it contains just the (primary) *when*-clause of `*start*`.

2. Repeat steps (a–c) until W = nil:

   (a) Unify the positive effects of all *when*-clauses in W with all possible operator preconditions, and mark the preconditions successfully matched in this way as "matched". (This marking is permanent.) Augment the *individual* domain of each matched precondition variable with a certain set C of constants, defined as follows. If the precondition variable was unified with a constant c, then C = {c}; if it was unified with an effect variable, then C is the *intersected* domain of that effect variable (relative to the *when*-clause to which the effect belongs).

   (b) Mark those *when*-clauses as "propagation candidates" that have all their preconditions (including corresponding primary preconditions) marked as "matched" and that involve at least one variable for which some relevant individual domain was augmented in step (a).

   (c) Reset W to nil. For all *when*-clauses that are propagation candidates, compute new intersected domains for their variables. If an intersected domain of a *when*-clause is thereby enlarged, and all intersected domains for the *when*-clause are now nonempty, then add the *when*-clause to W.

3. Further restrict intersected domains using equative preconditions of form (`EQ` u v), i.e., form a common intersected domain if both u and v are variables. If u is a constant and v is a variable, reduce the intersected domain of v by intersecting it with {u}; similarly if u is a variable and v is a constant. If the equation belongs to a primary *when*-clause, use it to reduce the intersected domains of u and v (whichever are variables) in the secondary clauses as well.

4. Return the intersected domains as the parameter domains, producing a sequence of lists with each list of form

   (*op* ($x_1$ $a_1$ $b_1$ $c_1$ ...) ($x_2$ $a_2$ $b_2$ $c_2$ ...) ...),

   where each operator *op* appears at least once. If *op* has $k$ conditional effects, there will be $k+1$ successive lists headed by *op*, where the first provides the parameter domains for the primary effects of *op* and the rest provide the parameter domains for the conditional effects (in the order of appearance in the UCPOP definition of *op*).

Note that we do not match or propagate *negative* conditions. The problem with negative conditions is that a very large number of them may be implicit in the initial conditions, given the use of the Closed World Assumption in UCPOP. For instance, in a world of $n$ blocks, with at most $O(n)$ on-relations (assuming that a block can be on only one other block), we necessarily have $O(n^2)$ implicit (not (on ...)) relations. In fact, the individual variable domains of negative preconditions or goals can really be infinitely large. For instance, given an empty initial state and a (paint-red ?x) operation with precondition (not (red ?x)) and effect (red ?x), we can achieve (red $c$) for infinitely many constants $c$. Perhaps negative conditions could be effectively dealt with by maintaining anti-domains for them, but we have not explored this since in practice ignoring negative conditions seems to cause only minimal "domain bloating". (We have proved that no actual domain elements can be *lost* through neglect of some preconditions.)

Our use of EQ-conditions could be refined by making use of them during the propagation process, and NEQ-conditions could also be used. However, doing so would probably have marginal impact.

As a final comment, we note that the output format specified in step 4 of the algorithm is actually generalized in our implementation so as to report inaccessible preconditions and goals. These inaccessible conditions are simply appended to the list of parameter domains for the appropriate *when*-clause of the appropriate operator. For instance, if the preconditions (oj ?oj) and (at ?oj ?city) in the ld-oj ("load orange juice") operator of the TRAINS world (see Online Appendix 2) are unreachable (say, because no oranges for producing orange juice have been provided), the parameter domain list for the (unique) *when*-clause of ld-oj will have the appearance

(ld-oj (?oj ...) (?car ...) (?city ...) (oj ?oj) (at ?oj ?city)).

This feature turns out to be very useful for debugging operator specifications and detecting unreachable goals.

### 4.3 Correctness and Tractability

In keeping with the remarks in the previous section, we will call an algorithm for computing parameter domains *correct* if the domains it computes subsume all possible parameter values that can actually occur (in a given primary or secondary *when*-clause) if we consider all possible sequences of operator applications starting at the given initial state.

The point is that this property will maintain the soundness of a planning algorithm that uses the precomputed parameter domains to prune impossible actions (as well as spurious threats) from a partially constructed plan. We assert the following:

**Theorem 1** *The* find-parameter-domains *algorithm is correct for computing parameter domains of* UCPOP*-style sets of operators (without quantification, disjunction, or facts), initial conditions, and (possibly) goal conditions.*

The proof is given in Appendix A. A preliminary step is to establish termination, using the monotonic growth of domains and the finiteness of the set of constants involved. Correctness is then established by showing that if there exists a valid sequence $A_0 A_1 ... A_n$ of actions (operator instances) starting with $A_0 = \text{*start*}$, and if $A_n$ is an instance of the operator Op, then the bindings that the parameters of Op received in instance $A_n$ are eventually added to the relevant intersected domains of Op (where "relevant" refers to the *when*-clauses of Op whose preconditions are satisfied at the beginning of $A_n$). This is proved by induction on $n$.

We now indicate how we can deal with universally quantified preconditions and effects, disjunctive preconditions, and facts. We make some simple changes to operator definitions by hand in preparation for parameter domain precomputation, and then use the domains computed by `find-parameter-domains`, together with the original operators, in running the planner. The steps for preparing an operator for parameter domain precomputation are as follows:

- Delete disjunctive preconditions, fact-preconditions,[13] and universally quantified preconditions (this includes universally quantified goals; it would also include universally quantified sentences embedded within the antecedents of *when*-clauses, e.g., in the manner (`:when` (`:forall` (`?x`) $\Phi$) $\Psi$), though these do not occur in any problem domains we have seen).

- Drop universal quantifiers occurring positively in operator effects, i.e., occurring at the top level or embedded by one or more `:and`'s. For example, an effect

  ```
  (:and (at robot ?to)
        (:not (at robot ?from))
        (:forall (?x)
             (:when (:and (grasping ?x) (object ?x))
                  (:and (at ?x ?to) (:not (at ?x ?from))) )))
  ```

  would become

  ```
  (:and (at robot ?to)
        (:not (at robot ?from))
        (:when (:and (grasping ?x) (object ?x))
             (:and (at ?x ?to) (:not (at ?x ?from))) ))
  ```

  Note that the universally quantified variable should be renamed, if necessary, to be distinct from all other such variables and from all operator parameters.

  In the example above the universally quantified variable is unrestricted. When the quantified variable includes a type restriction, as in (`:forall` (`object ?x`) $\Phi$), then this type restriction needs to become an antecedent of the matrix sentence $\Phi$. In the example at hand, $\Phi$ should be rewritten as the equivalent of (`:when` (`object ?x`) $\Phi$). Since $\Phi$ is often a *when*-clause, this can be done by adding (`object ?x`) as a conjunct to the antecedent of the *when*-clause. In some cases $\Phi$ is a conjunction of *when*-clauses, and in such a case the quantifier restriction can be added into each *when*-clause antecedent.

- Drop existential quantifiers in preconditions and goals, adding any restrictions on the quantified variables as conjuncts to the matrix sentence. For example, the goal

  ```
  (:exists (bacterium ?y)
           (:exists (molecule ?x)
                    (:and (contains IG ?x)
                          (contains ?x ?y)
                          (pure ?y) )))
  ```

---

13. E.g., in the STRIPS-world we would drop (`fact` (`loc-in-room ?x ?y ?room`)), which checks whether the given coordinates lie in the given room.

becomes

```
(:and (bacterium ?y) (molecule ?x) (contains IG ?x)
      (contains ?x ?y) (pure ?y) )
```

(Actually, the :and is dropped as well, when supplying goals to find-parameter-domains.)

With these reductions, find-parameter-domains will then compute correct parameter domains for the operators and goals. To see this, note first of all that dropping preconditions (in the initial step above) will not forfeit correctness, since doing so can only *weaken* the constraints on admissible parameter values, and thus can only *add* constants to the domains. The effect of dropping a universal quantifier, from the perspective of find-parameter-domains, is to introduce a new parameter in place of the universal variable. (The operator normalization subroutine detects variables in operator preconditions and effects that are not listed as parameters, and treats them as additional parameters.) While this is of course a drastic change in the meaning of the operator, it preserves correctness of the parameter domain calculation. This is because the domain of the new parameter will certainly contain all constants (and hence, under the Closed World Assumption, all objects) over which the quantified variable ranges. For example, if ?x is treated as a parameter rather than a universally quantified variable in the conditional effect

(:forall (?x) (:when (object ?x) (in ?x box))),

then the domain of ?x for the *when*-clause will consist of everything that can be an object, in any state where the operator can be applied. Thus the effect (in ?x box) will also be propagated for all such objects, as is required. Finally, the elimination of existential quantifiers from preconditions and goals can be seen to preserve the meaning of those preconditions and goals, and hence preserves the correctness of the parameter domain calculation.

Next we formally state our tractability claim for the algorithm, as follows (with some tacit assumptions, mentioned in the proof).

**Theorem 2** *Algorithm* find-parameter-domains *can be implemented to run in $O(mn_p n_e(n_p + n_e))$ time and $O(mn_p)$ space in the worst case, where $m$ is the number of constants in the problem specification, $n_p$ is the combined number of preconditions for all operators (and goals, if included), and $n_e$ is the combined number of operator effects (including those of* \*start\**).*

Again the proof is in Appendix A. The time complexity of find-parameter-domains is determined as the sum of (1) the cost of all the unifications performed, (2) the costs of all the individual domain updates attempted, and (3) the cost of all the intersected domain updates attempted. The space complexity bound is easily derived by assuming that there is a fixed upper bound on the number of arguments that a predicate (in a precondition or effect) can have, and from the fact that for each *when*-clause at most $O(m)$ constants are stored.

By adding some additional data structures in find-parameter-domains we can obtain a version of the algorithm whose worst-case time complexity is slightly improved. In fact, in step 2.(c) instead of propagating *all* of the effects of a *when*-clause with an enlarged intersected domain (i.e., adding such a *when*-clause to the list W), it is sufficient to propagate just those effects of the *when*-clause that involve an enlarged intersected-domain. This could be done by setting up for each *when*-clause a table that maps each parameter to a list of effects (of that *when*-clause) involving that parameter.

In the improved algorithm we use W to store the list of effects (instead of the list of *when*-clauses) that will be propagated in the next cycle of the algorithm, and steps 1, and 2 of `find-parameter-domains` are modified in the following way:

1'. Initialize W to the list of the effects of `*start*`.

2'. Repeat steps (a–c) until W = nil:

   (a') Unify the positive effects in W with all possible operator preconditions, and mark the preconditions successfully matched in this way as "matched" ...

   (b') same as 2.(b).

   (c') Reset W to nil. For all *when*-clauses that are propagation candidates, compute new intersected domains for their variables. If an intersected domain of a *when*-clause is thereby enlarged, and all intersected domains for the *when*-clause are now nonempty, then add to W the *subset* of the effects of the *when*-clause involving at least one parameter whose intersected domain is enlarged.

Note that the worst-case time complexity of the revised algorithm is improved, because now each effect of each *when*-clause is propagated at most $O(m)$ times. This decreases the upper bound on the number of unifications performed, reducing the complexity estimated in step (1) of the proof of Theorem 2 to $O(mn_e n_p)$. Hence we have proved the following corollary.

**Corollary 1** *There exists an improved version of* `find-parameter-domains` *that can be implemented to run in* $O(mn_p^2 n_e)$ *time in the worst case.*

## 5. Using Parameter Domains for Accelerating a Planner

We have already used the example of Molgen to motivate the use of precomputed parameter domains in planning, showing how such domains may allow us to prune non-viable actions from a partial plan.

More fundamentally, they can be used each time the planner needs to unify two predications involving a parameter, either during goal regression or during threat detection. (In either case, one predication is a (sub)goal and the other is an effect of an action or an initial condition.) If the unifier is inconsistent with a parameter domain, it should count as a failure even if it is consistent with other binding constraints in the current (partial) plan. And if there is no inconsistency, we can use the unifier *to intersect and thus refine the domains* of parameters equated by the unifier.

For example, suppose that G = (at ?x ?y) is a precondition of a step in the current plan, and that E = (at ?w ?z) is an effect of another (possibly new) step, where ?x, ?y, ?w and ?z are parameters (or, in the case of ?w and ?z, existentially quantified variables) which have no binding constraints associated with them in the current plan. Assume also that the domains of the parameters are:

```
?x : {Agent1, Agent2, Agent3}   ?y : {City1, City2}
?w : {Agent1, Agent2}           ?z : {City3, City4}
```

The unification of G and E gives the binding constraints {?x = ?w, ?y = ?z}, which are not viable because the parameter domains of ?y and of ?z have an empty intersection. On the other hand, if the domain of ?z had been {City2, City3, City4}, then the unification of G and E would have been judged viable, and the domains of the parameters would have been refined to:

```
?x : {Agent1, Agent2}    ?y : {City2}
?w : {Agent1, Agent2}    ?z : {City2}
```

Thus parameter domains can be incrementally refined as the planning search progresses; and the narrower they become, the more often they lead to pruning.

### 5.1 Incorporating Parameter Domains into UCPOP

The preceding consistency checks and domain refinements can be used in a partial-order, causal-link planner like UCPOP as follows. Given a goal (open condition) G selected by UCPOP as the next flaw to be repaired, we can

(1) restrict the set of new operator instances that UCPOP would use for establishing G; an instance of an operator with effect E (matching G) is disallowed if the precomputed parameter domains relevant to E are incompatible with the current parameter domains or binding constraints relevant to G; (note that the current parameter domains associated with G may be refinements of the initial domains);

(2) restrict the set of existing steps that UCPOP would reuse for establishing G; reusing a step with effect E (matching G) is disallowed if the current parameter domains relevant to E are incompatible with the current parameter domains or binding constraints relevant to G.

Moreover, given a *potential* threat by an effect Q against a protected condition P, inspection of the relevant parameter domains may reveal that the threat is actually spurious. This happens if the unifier of P and Q violates the (possibly refined) domain constraints of a parameter in P or Q. Thus we can often

(3) reduce the number of threats that are generated by the planner when a new causal link is introduced into the plan (this happens when an open condition is established either by reusing a step or by introducing a new one);

(4) recognize that a threat on the list of the flaws to be processed is redundant, allowing its elimination. (Note that since parameter domains are incrementally refined during planning, even if we use (3) during the generation of the threats, it is still possible for a threat to becomes spurious after it has been added to the flaw list).

These four uses of parameter domains cut down the search space without loss of viable solutions, since the options that are eliminated cannot lead to a correct, complete plan.

Note that (3) and (4) can be useful even when the planner only deals with definite threats (i.e., *d-sep* switch is turned on) for at least three reasons. First, determining that a threat is not a definite threat when *d-sep* is on incurs an overhead cost. So, earlier elimination of a spurious threat could lead to considerable savings if the threat is delayed many times during the search. The second reason relates to the plan-selection strategies adopted. If one uses a function that includes an (attenuated) term corresponding to the number of threats currently on the flaw list, then eliminating spurious threats in advance can give a more accurate measure of the "badness" of a plan. Finally, parameter domains could be used in threat processing so as to prune the search even when *dsep* is on. In particular, suppose that we modify the notion of a definite threat, when we have parameter domains, so that e.g., (P ?x) and (not (P ?y)) comprise a definite threat if the parameter domains associated with ?x and ?y are both c. So in that case, even with d-sep* on, we may discover early that a threat has become definite – in which case it might also be a forced threat, i.e., the choice between promotion and demotion may be dictated by ordering constraints; and that can prune the search space. However, in our current implementation we do not exploit this third point.

We have incorporated these techniques into UCPOP (version 2.0), along with our earlier improvements to the plan and goal selection strategies. Parameter domains are handled through an extension of the "VARSET" data structure (Weld, 1994) to include the domains of the variables (parameters), and by extending the unification process to implement the filtering discussed above.[14] We now describe our experiments with this enhanced system.

## 5.2 Experimental Results Using Parameter Domains

Our main goal here is to show that while the overhead determined by computing the parameter domains is not significant (both at preprocessing time and at planning time), exploitation of the parameter domains during planning can significantly prune the search. In the experiments we used the version of find-parameter-domains which is described in Section 4.2 and in Online Appendix 1. Note that for domains more complex than the ones we have considered it might be worthwhile to use the improved version of the algorithm discussed in Section 4.3. (However, it remains to be seen whether problems significantly more complex than those we consider here can be solved by any UCPOP-style planner.)

The CPU times needed by our implementation of find-parameter-domains are negligible for the problems we have looked at. They were 10 msec or less for many problems in the UCPOP test suite (when running compiled Allegro CL 4.2 on a SUN 20), 20 msec for two problems (Fixa from the fridge repair domain and Fixit from the flat tire domain), and 30msec on the TRAINS world problems described below.

In our first set of tests, we relied on the search strategy used as default in UCPOP. The function used for A* plan selection was thus S+OC+UC+F (allowing for problems that involve "facts"), and the goals were selected from the agenda according to a pure LIFO discipline.[15]

---

14. In the current implementation new threats are filtered only when the protected condition is established by a step already in the plan.
15. In all experiments the *d-sep* switch was on. The default delay-separation strategy for selecting unsafe conditions was slightly modified in the version of UCPOP using parameter domains. In particular, the

123

We began by experimenting with a variety of problems from UCPOP's test suite, comparing performance with and without the use of parameter domains. While relatively easy problems such as Sussman-anomaly, Fixa, Test-ferry, and Tower-invert4 showed no improvement through the use of parameter domains, most problems – particularly the harder ones – were solved more easily with parameter domains. For example, the Rat-insulin problem from the Molgen domain was solved nearly twice as fast, and some STRIPS-world problems (Move-boxes and variants)[16] and Towers of Hanoi (T-of-H1) were solved about 10 times as fast. Note that the STRIPS-world problems involve both facts and universally quantified conditional effects. Two problems from the office world, Office5 and Office6, which we knew to be readily solvable with our improved search strategy, remained difficult (in the case of Office6, unsolvable) with the default UCPOP strategy, despite the use of parameter domains.[17] Further experiments revealed that the source of this inefficiency was the default plan-selection strategy of UCPOP. In fact, using our S+OC+F strategy instead of S+OC+UC+F, without parameter domains Office5 and Office6 were solved generating/exploring 3058/2175 and 8770/6940 plans respectively; while using the parameter domains the plans numbered 1531/1055 and 2954/2204 respectively.

These initial experiments suggested to us that the most promising application of computed parameter domains would be for nontrivial problems that involved a variety of *types* of entities and relationships, and significant amounts of goal chaining (i.e., with each successive action establishing preconditions for the next). From this perspective, the TRAINS world struck us as a natural choice for further experimentation, with the additional advantage that its design was independently motivated by research at Rochester into mixed-initiative problem solving through natural-language interaction. (Refer again to the formalization in Online Appendix 2.) Recall from Table X that the Trains1 problem was extremely hard for unmodified UCPOP, requiring about 50 minutes and generating over a million plans.

Running the same problem with parameter domains produced a solution in 3.3 seconds (with 1207 plans generated), i.e., 927 times faster.

Intuitively, the use of parameter domains to constrain planning is analogous to using type constraints on the parameters (although parameter domains also take account of initial conditions). So it is of interest to see whether adding type constraints can provide similar efficiency gains as the use of parameter domains. Our first set of experiments therefore included T-Trains1, a "typed" version of Trains1; the operators have been slightly changed by adding new preconditions stating the types of the parameters involved. For example, the operator `uncouple` has been augmented with the preconditions (`engine ?eng`) and (`car ?car`). This problem was also extremely hard for the unmodified UCPOP, exceeding the search limit of 1,000,000 plans generated and requiring more than 2600 seconds. With parameter domains, the solution was obtained in one second.

---

    threats that can be resolved by separation and which are recognized to be redundant through the use of parameter domains were selected to be eliminated.

16. Move-boxes-2 differs slightly from the Move-boxes problem in the UCPOP suite, in that its goal is (`in-room box2 rclk`); Move-boxes-a differs slightly from the Move-boxes-2, in that its initial state contains two boxes.

17. Office5 is directly from UCPOP's test suite and Office6 is minor variant of Office5. In Office5, all persons are to be furnished with checks made out to them, using a check printer at the office and a briefcase for picking up the checks and bringing them home. "Sam" and "Sue" are the given persons, and in Office6 we have added (`person Alan`) and (`person Smith`) in the initial conditions.

| Problems | without domains | | with domains | | Domain ratio |
|---|---|---|---|---|---|
| | Plans | CPU sec | Plans | CPU sec | |
| Trains1 | 1,071,479/432,881 | 3050.15 | 1207/824 | 3.29 | 0.425 |
| T-Trains1 | > 1,000,000 | > 2335 | 404/296 | 0.98 | 0.425 |
| Move-boxes | 608,231/167,418 | 1024.04 | 5746/3253 | 18.8 | 0.705 |
| Move-boxes-1 | > 1,000,000 | > 6165 | 1264/645 | 3.59 | 0.705 |
| Move-boxes-2 | 13,816/3927 | 45.05 | 1175/587 | 2.66 | 0.705 |
| Move-boxes-a | 13,805/3918 | 46.11 | 1175/587 | 2.54 | 0.702 |
| T-of-H1 | 160,911/107,649 | 204.51* | 17,603/12,250 | 37.5 | 0.722 |
| Rat-insulin | 364/262 | 0.36 | 196/129 | 0.19 | 0.714 |
| Monkey-test1 | 96/62 | 0.12 | 75/46 | 0.11 | 0.733 |
| Monkey-test2 | 415/262 | 0.61 | 247/149 | 0.50 | 0.529 |
| Fix3 | 3395/2070 | 5.77 | 3103/1983 | 6.02 | 0.532 |
| Office5 | 809,345/500,578 | 1927.4 | 575,224/358,523 | 1556.8 | 0.625 |
| Office6 | > 1,000,000 | > 2730 | > 1,000,000 | > 2640 | 0.667 |
| Tower-invert4 | 806/538 | 1.55 | 806/538 | 1.59 | 0.733 |
| Sussman-anomaly | 44/26 | 0.05 | 44/26 | 0.06 | 0.917 |
| Fixa | 2131/1903 | 2.2 | 2131/1903 | 2.34 | 1 |
| Test-ferry | 718/457 | 0.65 | 718/457 | 0.71 | 1 |

Table XII: Plans generated/visited and CPU time (secs) for standard UCPOP with and without parameter domains. (* This result was obtained on a SUN 10 with Lucid Common Lisp; the others on a SUN 20 with Allegro Common Lisp.)

These results indicate that adding type constraints to operator specifications is not nearly as effective as the use of parameter domains in boosting planning efficiency. We discuss this point further in the context of the second set of tests (below).

Table XII summarizes the experimental results for all of the experiments that used the default UCPOP search strategy. The table gives the number of plans generated/visited by the planner and the CPU time (seconds) required to solve the problems.[18] Note that the use of the parameter domains gave very dramatic improvements not only in the TRAINS domain, but also in the STRIPS-world domain. The rightmost column supplies "domain ratio" data, as a metric that we hoped would predict the likely effectiveness of using parameter domains. The idea is that parameter domains should be effective to the extent that they filter out many parameter bindings that can be reached by chaining back from individual preconditions of an operator to the initial state. These bindings can be found by using a variant of the algorithm for propagating intersected domains that instead propagates *unions* of individual domains, and comparing these union domains to the intersected domains.[19]

---

18. The systems were compiled under Allegro CL 4.2, with settings (space 0) (speed 3) (safety 1) (debug 0), and run on a SUN 20. The CPU time includes the Lisp garbage collection (it is the time given in the output by UCPOP).

19. Actually, we do not need to explicitly propagate union domains, but can propagate (partial) bindings for one predication at a time, starting with the initial conditions. We match the predication to all possible preconditions, adding the constant arguments it contains to the union domains of the matched operator

| TRAINS problems | without domains | | with domains | | Domain ratio |
|---|---|---|---|---|---|
| | Plans | CPU sec | Plans | CPU sec | |
| Trains1 | 4097/2019 | 13.7 | 297/238 | 1.4 | 0.425 |
| Trains2 | 17,482/10,907 | 80.6 | 1312/1065 | 7.16 | 0.425 |
| Trains3 | 31,957/19,282 | 189.8 | 3885/3175 | 25.1 | 0.411 |

Table XIII: Plans generated/visited and CPU time (secs) for UCPOP with and without parameter domains in the TRAINS domain using the ZLIFO strategy.

| TRAINS problems | without domains | | with domains | | Domain ratio |
|---|---|---|---|---|---|
| | Plans | CPU sec | Plans | CPU sec | |
| Trains1 | 1093/597 | 8.1 | 265/194 | 2.3 | 0.425 |
| Trains2 | >50,000 | >607 | >50,000 | >534 | 0.425 |
| Trains3 | >50,000 | >655 | >50,000 | >564 | 0.411 |

Table XIV: Plans generated/visited and CPU time (secs) for UCPOP with and without parameter domains in the TRAINS domain using the LCFR strategy.

The "domain ratio" provides this comparison, dividing the average union domain size by the average intersected domain size, with averages taken over all parameters of all *when*-clauses of all operators.

The largest speedups (e.g., for the TRAINS problems) do tend to correlate with the smallest domain ratios, and the smallest speedups with the largest domain ratio (unity – see the last few rows). However, it can be seen from the table that the problem difficulty (as measured by plans or CPU time) is much more useful than the domain ratio as a predictor of speedups to be expected when using parameter domains. Problems that generate on the order of a million plans or more with standard UCPOP tend to produce speedups by 3 orders of magnitude, whereas the domain ratio for some of these problems (e.g., Move-boxes-1) is no better (or even worse) than for problems with much smaller speedups (e.g., Move-boxes-a, Rat-insulin, Monkey-test1, Monkey-test2). The much lower difficulty of these problems predicts their reduced speedup. But to complicate matters, not all difficult problems give high speedups (see T-of-H1 and especially Office5); we do not know what subtleties of problem structure account for these unusual cases.

In our second round of experiments, we tested the effectiveness of the parameter domain technique in combination with our improved search strategy, i.e., S+OC/ZLIFO. In addition, we combined S+OC with LCFR (least cost flaw selection) (Joslin & Pollack, 1994), so

---

(or *when*-clause). We then find corresponding (partially bound) effects, and add any new effects to the list of predications still to be propagated. A partially bound effect such as (P A ?x ?y) is new if there is no identical or similar predication such as (P A ?u ?v) among the previously propagated predications or among those still to be propagated.

| T-TRAINS | without domains | | with domains | | Domain |
| problems | Plans | CPU sec | Plans | CPU sec | ratio |
|---|---|---|---|---|---|
| T-Trains1 | 3134/2183 | 17.2 | 505/416 | 3.4 | 0.425 |
| T-Trains2 | 5739/4325 | 37.3 | 3482/2749 | 27.3 | 0.425 |
| T-Trains3 | 17,931/13,134 | 130.4 | 11,962/9401 | 105.1 | 0.425 |

Table XV: Plans generated/visited and CPU time (secs) for UCPOP with and without parameter domains in the "typed" TRAINS domain using the ZLIFO strategy.

| T-TRAINS | without domains | | with domains | | Domain |
| problems | Plans | CPU sec | Plans | CPU sec | ratio |
|---|---|---|---|---|---|
| T-Trains1 | 3138/2412 | 31.5 | 1429/1157 | 14.5 | 0.425 |
| T-Trains2 | >50,000 | >1035 | >50,000 | >1136 | 0.425 |
| T-Trains3 | >50,000 | >976 | >50,000 | >962 | 0.425 |

Table XVI: Plans generated/visited and CPU time (secs) for UCPOP with and without parameter domains in the "typed" TRAINS domain using the LCFR strategy.

as to test for possible sensitivity of the parameter-domains technique to the precise strategy used. For the present set of tests we used a search limit of 50,000 plans generated.

Once again we began by sampling some problems from the UCPOP test suite, and these initial trials yielded results quite analogous to those for the default UCPOP strategy. We obtained no improvements for several easier problems and significant improvements for harder ones (e.g., again close to a factor of 2 for Rat-insulin). Noteworthy members of the latter category were Office5 and Office6 – recall that Office5 had shown little speedup with standard UCPOP and Office6 had been unsolvable. However, in view of the computational expense of testing both ZLIFO and LCFR, we then decided to narrow our focus to the TRAINS world. As mentioned, the advantages of this world are its inherent interest and relative complexity.

Tables XIII-XVI provide experimental results for the TRAINS domain with the S+OC/ZLIFO strategy and the S+OC/LCFR strategy, in each case with and without parameter domains.

The results in Tables XIII and XIV show that using parameter domains can still give very significant improvements in performance, over and above those obtained through the use of better search strategies. For example, the use of parameter domains provided an 11-fold speedup for Trains2, for the S+OC/ZLIFO strategy. In this particular problem the speedup (on all metrics) was the result of pruning 1482 plans (more than half of those generated) during the search., and recognizing 305 unsafe conditions as redundant. Evidently, the effect of this pruning is amplified by an order of magnitude in the overall performance, because of the futile searches that are cut short. Note that the speedups for Trains1-3 are roughly comparable (within a factor of 2) to those obtained for problems in the previous set with comparable initial difficulty (e.g., see Move-boxes-2 and Move-boxes-a in Table XII). This again points to a rather consistent correlation between problem difficulty and speedups obtainable using parameter domains. The constant domain ratios are also compatible with the more or less invariant speedups here, though this is of little import, given the earlier results. For S+OC/LCFR the gains appear to be less, though the single result showing a 3.5-fold speedup provides only anecdotal evidence for such a conclusion. Trains2 and Trains3 remained too difficult for solution by LCFR. Similar gains were observed for the S+OC/LC strategies where the best observed gain in the Trains domain was a 1.7-fold speedup for Trains2. In any case, all results confirm the effectiveness of the parameter-domains technique.

Tables XV and XVI are again for the "typed" version of TRAINS. In this case parameter typing gave modest improvements in the absence of parameter-domains, and (in contrast with the results for Trains1 under the default search strategy) significant deterioration in their presence. While we do not know how to account for these results in detail, it seems clear that contrary effects are involved. On the one hand, typing does tend to help in that it tends to limit choices of parameter values to "sensible" ones. For example, a precondition (`engine ?eng`) will be satisfiable only through use of `*start*`, and the initial state will thus constrain `?eng` to assume sensible values. On the other hand, adding type-preconditions will tend to broaden the search space, by adding further open conditions to the flaw list.

The lesson from the "typed" experiments appears to be that it is best *not* to supply explicit type constraints on operator parameters, instead using our automated method of calculating and updating domains to constrain parameter bindings.

## 6. Conclusions and Further Work

We began by exploring some simple, domain-independent improvements to search strategies in partial order planning, and then described a method of using precomputed parameter domains to prune the search space. We now summarize our conclusions about these techniques and then point to promising directions for further work.

### 6.1 Improving Search

Our proposed improvements to search strategies were based on the one hand on a carefully considered choice of terms in the A* heuristic for plan selection, and on the other on a preference for choosing open conditions that cannot be achieved at all or can be achieved in only one way (with a default LIFO prioritization of other open conditions). Since the plan refinements corresponding to uniquely achievable goals are logically necessary, we have termed the latter strategy a zero-commitment strategy. One advantage of this technique over other similar strategies is that it incurs a lower computational overhead.

Our experiments based on modifications of UCPOP indicate that our strategies can give large improvements in planning performance, especially for problems that are hard for UCPOP (and its "relatives") to begin with. The best performance was achieved when our strategies for plan selection and goal selection were used in combination. In practical terms, we were able to solve nearly every problem we tried from the UCPOP test suite in a fraction of a second (except for Fixit, which required 38.2 seconds), where some of these problems previously required minutes or were unsolvable on the same machine. This included a sufficient variety of problems to indicate that our techniques are of broad potential utility.

Further, our results suggest that zero-commitment is best supplemented with a LIFO strategy for open conditions achievable in multiple ways, rather than a generalization of zero-commitment favoring goals with the fewest children. This somewhat surprising result might be thought to be due to the way in which the designer of a domain orders the preconditions of operators; i.e., the "natural" ordering of preconditions may correlate with the best planning order, giving a fortuitous advantage to a LIFO strategy relative to a strategy like LC.[20]

However, some preliminary experiments we performed with randomized preconditions for T-of-H1 and Trains1 indicate otherwise. In 5 randomizations of the preconditions of T-of-H1, both LC and ZLIFO were slowed down somewhat, by average factors of 2.2 (2) and 3.3 (4.2) in terms of plans expanded (CPU time used) respectively. (In both cases, S+OC was used for plan search.) This still left ZLIFO with a performance advantage of a factor of 22 in terms of plans created and 39 in terms of CPU time. For Trains1 the performance of LC greatly deteriorated in 2 out of 5 cases (by a factor close to 70 in terms of both plans and time), while that of ZLIFO actually improved marginally. This now left ZLIFO with an average performance advantage over LC (whereas it had been slightly slower in the unrandomized case) — a factor of 3.3 in terms of plans and 6.7 in terms of CPU time (though these values are very unreliable, in view of the fact that the standard deviations are of the same order as the means).

Despite these results we believe that a satisfactory understanding of the dependence of flaw-selection strategies on the order of operator preconditions will require a more extensive experimental investigation. We are currently undertaking this work.

### 6.2 Using Parameter Domains

We described an implemented, tractable algorithm for precomputing parameter domains of planning operators, relative to given initial conditions. We showed how to use the precomputed domains during the planning process to prune non-viable actions and bogus threats, and how to update them dynamically for maximum effect.

The idea of using precomputed parameter domains to constrain planning was apparently first proposed in a technical report by Goldszmidt et al. (1994). This contains the essential idea of accumulating domains by forward propagation from the initial conditions. Though the report only sketches a single-sweep propagation process from the initial conditions to the goals, the implemented Rockwell Planner (RNLP) handles cyclic operator graphs, repeatedly propagating bindings until quiescence, much as in our algorithm. Our algorithm deals with the additional complexities of conditional effects and equalities (and in semi-automated fashion with quantification) and appears to be more efficient (Smith, 1996). Other distinctive features of our work are the method of incrementally refining domains

---

20. This was suggested to us by David Smith as well as Mike Williamson. Williamson tried ZLIFO with 5 randomized versions of T-of-H1, and reported a large performance degradation (Williamson & Hanks, 1996). We recently ran these versions using our implementation, obtaining far more favorable results (three of the five versions were easier to solve than the original version of T-of-H1, while the other two versions slowed down ZLIFO by a factor of 1.84 and 4.86 in terms of plans explored.)

during planning, the theoretical analysis of our algorithm, and the systematic experimental tests.

Another closely related study is that of Yang and Chan (1994), who used hand-supplied parameter domains in planning much as we use precomputed domains. An interesting aspect of their work is the direct use of sets of constants as variable bindings. For instance, in establishing a precondition (P ?x) using an initial state containing (P a), (P b) and (P c), they would bind ?x to {a, b, c} rather than to a specific constant. They refine these "noncommittal" bindings during planning much as we refine variable domains, and periodically use constraint satisfaction methods to check their consistency with current EQ/NEQ constraints. They conclude that delaying variable bindings works best for problems with low solution densities (while degrading performance for some problems with high solution densities), and that the optimal frequency of making consistency checks depends on whether dead ends tend to occur high or low in the search tree. Our work is distinguished from theirs by our method of precomputing parameter domains, our use of specific bindings when matching initial conditions to OCs, our use of parameter domains in threat detection and resolution, and our handling of the enriched syntax of UCPOP operators as compared SNLP operators.

Judging from the examples we have experimented with, our techniques are well-suited to nontrivial problems that involve diverse types of objects, relations and actions, and significant logical interdependencies among the steps needed to solve a problem. When used in conjunction with the default search strategy of UCPOP, our method gave significant speedups for nontrivial problems, reaching a speedup factor of 927 in the TRAINS transportation planning domain, and more than 1717 for the hardest STRIPS-world problem we tried . When combined with our S+OC and ZLIFO search strategies, the parameter domain technique still gave speedups by a factor of around 10 for some TRAINS problems. Though our implementation is aimed at a UCPOP-style planner, essentially the same techniques would be applicable to many other planners.

We also found the parameter domain precomputations to be a very useful debugging aid. In fact, the domain precomputation for our initial formulation of the TRAINS world immediately revealed several errors. For instance, the domain of the ?eng parameter of mv-engine turned out to contain oranges, bananas, and an OJ factory, indicating the need for a type constraint on ?eng. (Without this, transportation problems would have been solvable without the benefit of engines and trains!) Another immediately apparent problem was revealed by the parameter domains for ?city1 and ?city2 in mv-engine: the domain for ?city1 excluded Elmira, and that for ?city2 excluded Avon. The obvious diagnosis was that we had neglected to assert both  (connected c1 c2) and (connected c2 c1) for each track connecting two cities. Furthermore, the parameter domains can quickly identify unreachable operators and goals in some cases. For instance, without the make-oj operator, the computed domains show that the ld-oj operator is unreachable, and that a goal like (and (oj ?oj) (at ?oj Bath)) (getting some orange juice to Bath) is unattainable (the parameter domain for ?oj will be empty).

Of course, running the planner itself can also be used for debugging a formalization, but planning is in general far more time-consuming than our form of preprocessing (especially if the goal we pose happens to be unachievable in the formalization!), and the trace of an anomalous planning attempt can be quite hard to interpret, compared to a listing of parameter domains, obtained in a fraction of a second.

### 6.3 Further work

First of all, some additional experimentation would be of interest, to further assess and perhaps refine our search strategies. Some of this experimentation might focus on threat-handling strategies, including the best general form of an attenuated UC-term in plan selection, and the best way to combine threat selection with open condition selection. The preference for definite threats over open conditions used by ZLIFO does appear to be a good default according to our experience, but the TileWorld experiments indicated that a re-ordering of priorities between threats and open conditions is sometimes desirable. Concerning the choice of a UC-related term for inclusion in the heuristic for plan selection, we should mention that we have briefly tried using $S+OC+UC_d$, where $UC_d$ is the number of definite threats, but did not obtain significant uniform improvements.

One promising direction for further development of our search strategy is to make the zero-commitment strategy apply more often by finding ways of identifying false options as early as possible. That is, if a possible action instance (obtained by matching an open condition against available operators as well as against existing actions) is easily recognizable as inconsistent with the current plan, then its elimination may leave us with a single remaining match and hence an opportunity to apply the zero-commitment strategy.

One way of implementing this strategy would be to check at once, before accepting a matched action as a possible way to attain an open condition, whether the temporal constraints on that action force it to violate a causal link, or alternatively, force its causal link to be violated. In that case the action could immediately be eliminated, perhaps leaving only one (or even no) alternative. This could perhaps be made even more effective by broadening the definition of threats so that preconditions as well as effects of actions can threaten causal links, and hence bring to light inconsistencies sooner. Note that if a precondition of an action is inconsistent with a causal link, it will have to be established with another action whose *effects* violate the causal link; so the precondition really poses a threat from the outset.

Two possible extensions to our parameter domain techniques are (i) fully automated handling of universally quantified preconditions and effects, disjunctions and facts in the preprocessing algorithm; and (ii) more "intelligent" calculation of domains, by applying a constraint propagation process to the sets of ground predications that have been matched to the preconditions of an operator; this can be shown to yield tighter domains, though at some computational expense. Blum and Furst (1995) recently explored a similar idea, but rather than computing parameter domains, they directly stored sets of ground atoms that could be generated by one operator application (starting in the initial state), two successive operator applications, and so on, and then used these sets of atoms (and exclusivity relations among the atoms and the actions connecting them) to guide the regressive search for a plan. The algorithm they describe does not allow for conditional effects, though this generalization appears entirely possible. For the examples used in their tests, they obtained dramatic speedups.

Finally, we are also working on another preprocessing technique, namely the inference of state constraints from operator specifications. One useful form of constraint is implicational (e.g., (implies (on ?x ?y) (not (clear ?y)))), and another is single-valuedness conditions (e.g., (on ?x ?y) may be single-valued in both ?x and ?y). We conjecture that such constraints can be tractably inferred and used for further large speedups in domain-independent, well-founded planning.

In view of the results we have presented and the possibilities for further speedups we have mentioned, we think it plausible that well-founded, domain-independent planners may yet become competitive with more pragmatically designed planners.

## Acknowledgements

This work amalgamates and extends two conference papers on improving search (Schubert & Gerevini, 1995) and using computed parameter domains (Gerevini & Schubert, 1996) to accelerate partial-order planners. The research was supported in part by Rome Lab contract F30602-91-C-0010 and NATO Collaborative Research Grant CRG951285. Some of the work by AG was carried out at IRST, 38050 Povo (TN), Italy, and at the CS Department of the University of Rochester, Rochester NY USA. The helpful comments and perceptive questions of Marc Friedman, David Joslin, Rao Kambhampati, Colm O'Riain, Martha Pollack, David Smith, Dan Weld, Mike Williamson, and of Associate Editor Michael Wellman and the anonymous reviewers are gratefully acknowledged.

**Appendix A** (Proofs of the Theorems)

**Theorem 1** *The* find-parameter-domains *algorithm is correct for computing parameter domains of* UCPOP*-style sets of operators (without quantification, disjunction, or facts), initial conditions, and (possibly) goal conditions.*

*Proof.* As a preliminary observation, the intersected parameter domains computed iteratively by the algorithm eventually stabilize, since they grow monotonically and there are only finitely many constants that occur in the initial conditions and in operator effects. Thus the algorithm terminates.

In order to prove correctness we need to show that if there exists a valid sequence $A_0 A_1 ... A_n$ of actions (operator instances) starting with $A_0 = $ *start*, and if $A_n$ is an instance of the operator Op, then the bindings that the parameters of Op received in instance $A_n$ are eventually added to the relevant intersected domains of Op (where "relevant" refers to the *when*-clauses of Op whose preconditions are satisfied at the beginning of $A_n$). We prove this by induction on $n$.

If $n = 0$, then $A_n = A_0 = $ *start*, so there are no parameters and the claim is trivially true.

Now assume that the claim holds for $n = 1, 2, ..., k$. Then consider any operator instance $A_{k+1}$ that can validly follow $A_0 A_1 ... A_k$, i.e., such that $A_{k+1}$ is an instance of an operator Op whose primary preconditions, possibly along with some secondary ones, are satisfied at the end of $A_0 A_1 ... A_k$. Let $p$ be such a precondition, and write its instance in $A_{k+1}$ as (P c1 c2 ..). Then (P c1 c2 ..) must be an effect of some $A_i$, where $0 \leq i \leq k$. If $i = 0$ then (P c1 c2 ..) holds in the initial state, and hence this predication is propagated and successfully matched to $p$ in the initial propagation phase of `find-parameter-domains`. If $i > 0$, then $A_i$ is an instance of some operator Op' and (P c1 c2 ..) is the corresponding instance of some effect (P $t_1$ $t_2$ ..) of Op', where each $t_j$ is either a parameter of Op' or is equal to cj. Diagrammatically,

$$A_0 \quad \ldots \quad A_i \quad \ldots \quad A_k \quad A_{k+1}$$

$$\begin{array}{ccc} \text{Op'} & & \text{Op} \\ \text{effect (P } t_1\ t_2\ ..) & \longrightarrow & \text{precond } p \\ \text{(P c1 c2 ..)} & & \text{(P c1 c2 ..)} \end{array}$$

By the induction assumption, the bindings of the parameters in $A_i$ are eventually added to the relevant intersected domains of Op'. This also implies that the intersected domains of Op' become nonempty, and so the effect (P $t_1$ $t_2$ ..) is eventually propagated, where any variables among the $t_j$ have the corresponding constant cj in the relevant intersected domain. Consequently, much as in the case $i = 0$, effect (P $t_1$ $t_2$ ..) is successfully matched to precondition $p$ of Op at some stage of the propagation. Given these observations, it is clear that for both $i = 0$ and $i > 0$, $p$ will be marked "matched" in Op eventually, and furthermore any parameters of Op that occur in $p$ will have the bindings resulting from the unification with (P c1 c2 ..) added to the appropriate *individual* domains associated with $p$.

This argument applies to all preconditions of Op satisfied in its instance $A_{k+1}$, in particular to all the primary preconditions. Since these are all marked "matched", the algorithm will compute intersected domains for all Op-parameters that occur in them. In view of the individual domain updates just confirmed, and since individual domains grow monotonically, these intersected domains will eventually contain the parameter bindings of $A_{k+1}$. For instance, if a parameter ?x of Op occurs in a primary precondition and is bound to c in $A_{k+1}$, we have shown that c will eventually be added to the intersected domain of ?x associated with the primary *when*-clause of Op. If a parameter does not occur in the primary preconditions of Op, then its intersected domain is set to T at the outset, and this implicitly contains whatever binding the parameter has in $A_{k+1}$.

A very similar argument can be made for any secondary *when*-clause of Op whose preconditions are also satisfied for $A_{k+1}$. Again, all preconditions of the secondary clause, as well as the primary preconditions, will be marked "matched", and so for any parameter occurring in these combined preconditions, its intersected domain (relative to the secondary clause) will be updated to include its binding in $A_{k+1}$. For parameters of Op not occurring in any of these preconditions, the intersected domains will again be set to T initially, and this implicitly contains any possible binding. Finally, we note that since the intersected domains relative to both primary and secondary *when*-clauses grow monotonically, the augmentations of intersected domains we have just confirmed is permanent. (In the case of T-domains, these remain T.)

We leave some additional details concerned with the ultimate use of EQ-preconditions in `find-parameter-domains` to the reader. □

**Theorem 2** *Algorithm* `find-parameter-domains` *can be implemented to run in* $O(mn_p n_e(n_p + n_e))$ *time and* $O(mn_p)$ *space in the worst case, where* $m$ *is the number of constants in the problem specification,* $n_p$ *is the combined number of preconditions for all operators (and goals, if included), and* $n_e$ *is the combined number of operator effects (including those of* `*start*`*).*

*Proof.* The time complexity of `find-parameter-domains` can be determined as the sum of (1) the cost of all the unifications performed, (2) the costs of all the individual domain updates attempted, and (3) the cost of all the intersected domain updates attempted. We estimate an upper bound for each of these terms under the following assumptions:

(a) the unification of any operator effect with any operator precondition requires constant time;

(b) there is a fixed upper bound on the number of arguments that a predicate (in a precondition or effect) can have. It follows that $O(n_e)$ is an upper bound on the total number of intersected domains;[21]

(c) individual domains and intersected domains are stored in hash tables (indexed by the constants in the domain). So, we can check whether an element belongs to a particular (individual or intersected) domain, and possibly add it to that domain essentially in constant time. Furthermore for each individual and intersected domain, appropriate data structures are used to keep track of the (possibly empty) set of new elements that have been added to the domain in the last update attempt.

(1) For any particular intersected domain of any particular operator, there can be at most $m$ updates of this domain. Each such update causes all of the effects of the *when*-clause to which the intersected domain belongs to be propagated. An upper bound on this number is $n_e$. Each propagated effect may then be unified with $O(n_p)$ preconditions. Thus the $O(m)$ updates of an intersected domain may cause $O(mn_e n_p)$ unifications. Hence from (b), the overall number of unifications caused by the propagation of intersected domains to individual domains is $O(mn_e^2 n_p)$. To these unifications we have to add those which are initially performed between the effects of `*start*` and the preconditions of the operators. There are $O(mn_p)$ such unifications, and so they do not increase the previous upper bound on the number of unifications. Thus, from (a), the cost of all of the unifications performed by the algorithm is $O(mn_e^2 n_p)$.

(2) Each unification is potentially followed by an attempt to update the individual domain(s) of the relevant parameter(s). However, with assumption (c) the number of such attempts is limited to those where the set of *new* elements in the intersected domain(s) of the unifying effect is (are) *not* empty. Furthermore, when we attempt to update an individual domain $D_I$ by performing the union of a relevant intersected domain $D_i$ and $D_I$, only the subset of the new elements of $D_i$ need to be added to $D_I$ (if they are not already there). Thus, since any intersected domain grows monotonically, from (b) and (c) we have that the overall cost of all the update attempts for one particular individual domain caused by one particular effect is $O(m)$. But in the worst case one effect can unify with all the $O(n_p)$ preconditions of all the operators, yielding an overall bound on all of the attempts to update the individual domains of $O(mn_e n_p)$.

---

21. Note that if a parameter appears in a precondition of a *when*-clause, but in none of its effects, then the intersected domain of the parameter will not be propagated by the algorithm. Hence in implementing the algorithm we can ignore such parameters.

(3) There can be an attempt to update a particular intersected domain for each relevant individual domain update, and each relevant individual domain can be updated $O(m)$ times (because the domains grow monotonically). Therefore from (b) there are at most $O(mn_p)$ attempts to update one intersected domain. By (c) the total cost of these attempts is $O(mn_p^2)$, because checking whether a *new* element of an individual domain belongs to all the other $O(n_p)$ relevant individual domains takes $O(n_p)$ time. So, since from (b) there are no more than $O(n_e)$ intersected domains, the total cost incurred by the algorithm for updating all of the intersected domains is $O(mn_e n_p^2)$.

It follows that the time complexity of `find-parameter-domains` is:
$$O(mn_e^2 n_p) + O(mn_e n_p) + O(mn_e n_p^2) = O(mn_p n_e(n_p + n_e)).$$

The space complexity bound is easily derived from (b), and from the fact that for each *when*-clause at most $O(m)$ constants are stored. □

Accelerating Partial-Order PlannersSrinivasan, R., & Howe, A. (1995). Comparison of methods for improving search efficiency in a partial-order planner. In *Proceedings of the Fourteenth International Joint Conference on Artificial Intelligence (IJCAI-95)*, pp. 1620–1626.

Weld, D. (1994). An introduction to least commitment planning. *AI Magazine*, *15*(4), 27–62.

Wilkins, D. (1988). *Practical Planning: Extending the Classical AI Planning Paradigm.* Morgan Kaufmann, San Mateo, CA.

Williamson, M., & Hanks, S. (1995). Flaw selection strategies for value-directed planning. In *Proceedings of the Third International Conference on Artificial Intelligence Planning Systems*, pp. 237–244.

Yang, Q., & Chan, A.Y.M. (1994). Delaying variable binding commitments in planning. In *Proceedings of the Second International Conference on Artificial Intelligence Planning Systems*, pp. 182–187.137